\documentclass{article}
\pdfoutput=1
\usepackage[numbers,sort&compress]{natbib}

\usepackage{arxiv}

\usepackage[utf8]{inputenc} 
\usepackage[T1]{fontenc}    
\usepackage{hyperref}       
\usepackage{url}            
\usepackage{booktabs}       
\usepackage{amsfonts}       
\usepackage{nicefrac}       
\usepackage{microtype}      
\usepackage{lipsum}
\usepackage{graphicx}
\graphicspath{ {./images/} }

\usepackage[frozencache, cachedir=.]{minted}
\usepackage{booktabs}
\usepackage{xcolor}
\usepackage{colortbl}
\usepackage{listings}
\usepackage{hyperref}

\usepackage{amsmath} 
\usepackage{amssymb}

\definecolor{bg}{rgb}{0.95,0.95,0.95}

\usepackage{blindtext}
\usepackage{geometry}
\usepackage{float}
\usepackage[most]{tcolorbox}
\usepackage{tikz}
\usepackage{varwidth}
\usepackage{capt-of}

\usepackage{soul} 

\tcbset{enhanced,colback=cyan!5!white,colframe=cyan!75!black,fonttitle=\bfseries}

\newcommand{\lib}{\textit{Kornia}}

\title{A survey on \lib: an Open Source Differentiable Computer Vision Library for PyTorch}

\author{
 Edgar Riba \\
  Computer Vision Center, Universitat Autònoma de Barcelona \\
  Institut de Robòtica i Informàtica Industrial, Barcelona \\
  OpenCV.org \\
  \texttt{edgar.riba@gmail.com} \\
   \And
 Dmytro Mishkin \\
  Faculty of Electrical Engineering, \\
  Czech Technical University in Prague \\
  \texttt{ducha.aiki@gmail.com} \\
  \And
 Jian Shi \\
  Department of Ophthalmology and Visual Sciences, \\
  The Chinese University of Hong Kong \\
  \texttt{jianshi@cuhk.edu.hk} \\
  \And
 Daniel Ponsa \\
  Computer Vision Center, \\
  Universitat Autònoma de Barcelona \\
  \texttt{daniel@cvc.uab.es} \\
  \And
 Francesc Moreno-Noguer \\
  Institut de Robòtica i Informàtica Industrial, Barcelona \\
  \texttt{fmoreno@iri.upc.edu} \\
  \And
  Gary Bradski \\
  OpenCV.org \\
  \texttt{garybradski@gmail.com} \\
}

\begin{document}
\maketitle
\begin{abstract}
This work presents \lib{}, an open source computer vision library built upon a set of differentiable routines and modules that aims to solve generic computer vision problems. The package uses \textit{PyTorch} as its main backend, not only for efficiency but also to take advantage of the reverse auto-differentiation engine to define and compute the gradient of complex functions. Inspired by \textit{OpenCV}, \lib{} is composed of a set of modules containing operators that can be integrated into neural networks to train models to perform a wide range of operations including image transformations, camera calibration, epipolar geometry, and low level image processing techniques, such as filtering and edge detection that operate directly on high dimensional tensor representations on graphical processing units, generating faster systems. Examples of classical vision problems implemented using our framework are provided including a benchmark comparing to existing vision libraries.
\end{abstract}


\section{Introduction}

\begin{figure}[t]
    \setlength\tabcolsep{2.5pt}
    \begin{center}
        \begin{tabular}{c c c c c}
        \textbf{Color} & \textbf{Enhancement} & \textbf{Filtering} & \textbf{Features} & \textbf{Geometry}\\
        \includegraphics[width=2.5cm]{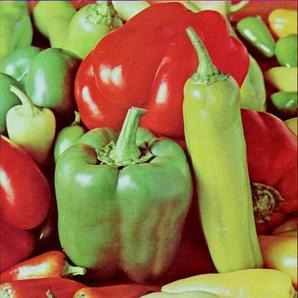} &
        \includegraphics[width=2.5cm]{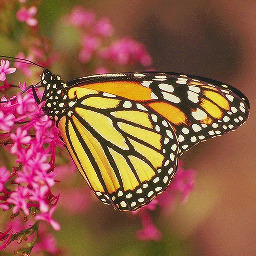} &
        \includegraphics[width=2.5cm]{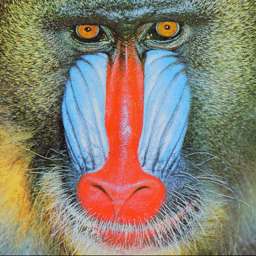} &
        \includegraphics[width=2.5cm]{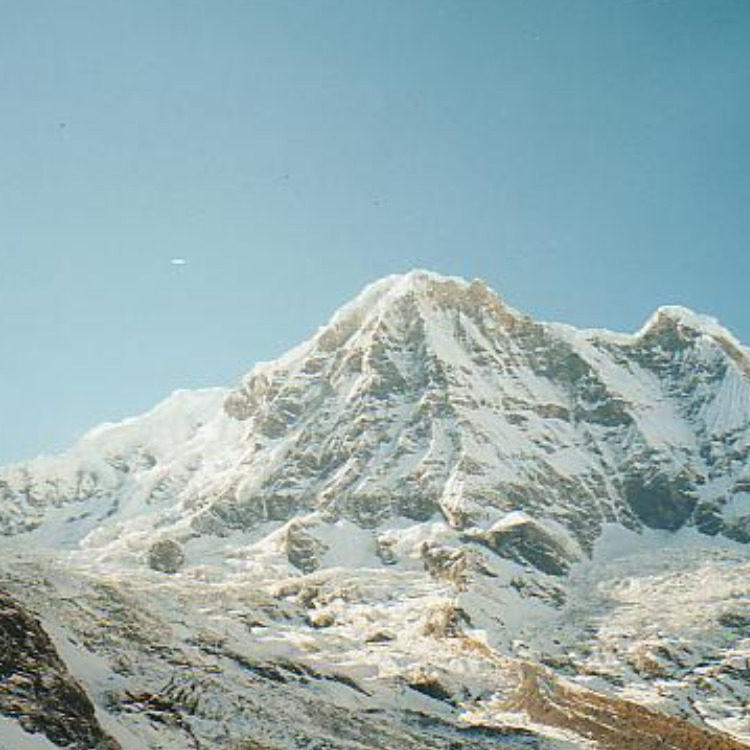} &
        \includegraphics[width=2.5cm]{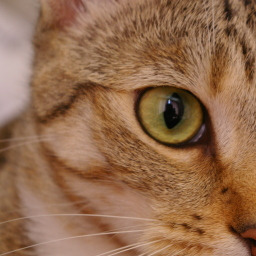} \\
        \includegraphics[width=2.5cm]{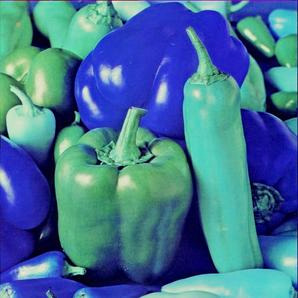} &
        \includegraphics[width=2.5cm]{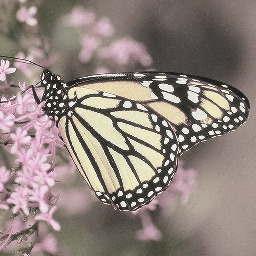} &
        \includegraphics[width=2.5cm]{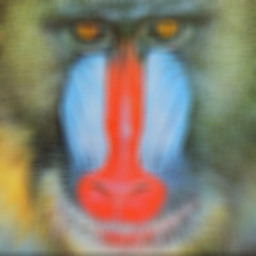} &
        \includegraphics[width=2.5cm]{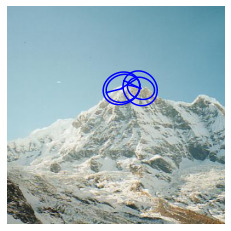} &
        \includegraphics[width=2.5cm]{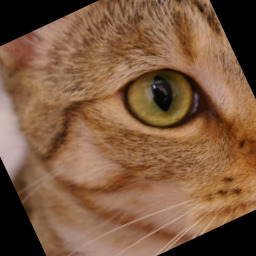} \\
        \includegraphics[width=2.5cm]{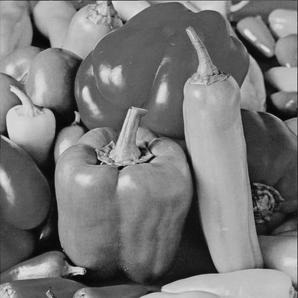} &
        \includegraphics[width=2.5cm]{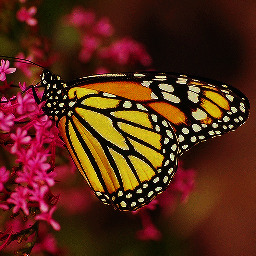} &
        \includegraphics[width=2.5cm]{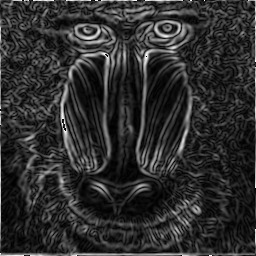} &
        \includegraphics[width=2.5cm]{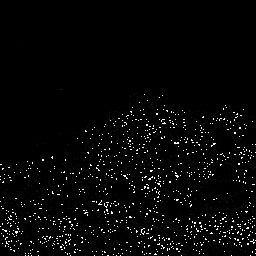} &
        \includegraphics[width=2.5cm]{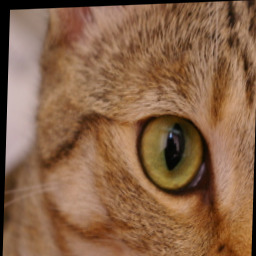} \\
        \end{tabular}
    \end{center}
    \caption{\lib{} implements routines for low level image processing tasks using native \textit{PyTorch} operators and taking advantage of its high-performance optimizations. The purpose of the library is to be used as a base for large-scale vision projects, data augmentation frameworks, or for creating computer vision layers inside of neural network layers that allow for back-propagating the error through them. The results in this figure are obtained from a given batch of image tensor using data parallelism in the GPU. More examples showing the usage of the library on this specific tasks plus other related to vision are provided in the \underline{\color{blue}\href{https://github.com/kornia/kornia-examples}{kornia-examples}} repository.}
    \label{fig:imgproc}
\end{figure}

Computer vision has driven a lot of technological advances in modern society for many different industries such as Automotion to improve the perception algorithms for self-driving cars; Factory Automation precisely in robotics field; or Audio Visual Production for visual effects generation. One of the key components of this achievements has been thanks to open source software that provided to the community free and accessible implementations of the main computer vision and machine learning algorithms.

There exist several open-source libraries widely used by the computer vision community which are tailored to process images using Central Processing Units (CPUs) such as OpenCV~\citep{opencv}, scikit-image~\citep{scikit-image}, or Pillow~\citep{pillow} and many others that are optimised for specific use cases. However, nowadays many of the top performing computer vision algorithms rely on deep learning models, with the huge need to process images in parallel using Graphical Processing Units (GPUs) in order to achieve high-performance requirements. Within that context, during the last couple of years many frameworks for deep learning have gained a lot popularity; to mention some of them: PyTorch~\citep{pytorch}, Tensorflow~\citep{tensorflow2015-whitepaper}, Caffe~\citep{caffe}, MXNet~\citep{journals/corr/ChenLLLWWXXZZ15}, or MatConvNet~\citep{Vedaldi15}. In concrete, PyTorch \citep{pytorch} due to its reverse-mode automatic differentiation engine, dynamic computation graph, distributed learning, eager/script execution modes and its intuitive API introduces a different paradigm within the community. PyTorch and its ecosystem provide a few packages to work with images such as its most popular toolkit, \textit{torchvision}, which is mainly designed to perform data augmentation, read popular datasets and implementations of state-of-the-art models for tasks such as detection, segmentation, image generation, and landmark detection. Despite all these virtues, PyTorch is still lacking in implementations for classical vision algorithms using their native tensor data structures and making them efficient to be used on GPUs or any high-permanence device supported by PyTorch.

This paper introduces \lib, an open source computer vision library built on top of PyTorch that is intended to help students, researchers, companies and entrepreneurs to implement computer vision applications oriented towards deep learning. Our library, in contrast to standard vision frameworks, provides classical and advanced image processing algorithms implemented such that they can be embedded into deep networks. \lib{} is designed to fill the gap between PyTorch and computer vision communities and it is based on some of the pre-existing open source solutions for computer vision (Pillow Image (PIL), skimage, torchvision, Tensorflow.image), but with a strong inspiration on OpenCV~\citep{opencv}. As shown in Table~\ref{tab:table-cv-frameworks}, \lib{}, in contrast to other existing libraries, which are limited to its usage in CPU and similar to tf.image making use of differentiability on the GPU, tries to combine the simplicity of OpenCV and PyTorch in order to leverage differentiable programming for computer vision  borrowing some properties from PyTorch such as differentiability, GPU acceleration, distributed data-flow and production quality code.

\begin{table*}[t]
\centering
\begin{tabular}{l*{6}{c}r}
                   & CPU & GPU & Batched & Differentiable & Distributed & ND-Array \\
\hline
scikit-image & \checkmark & $\times$ & $\times$ & $\times$ & $\times$ & $\times$ \\
Numpy and scipy  & \checkmark & $\times$ & $\times$ & $\times$ & $\times$ & $\times$ \\
Albumentations & \checkmark & $\times$ & $\times$ & $\times$ & $\times$ & $\times$ \\
torchvision  & \checkmark & $\times$ & $\times$ & $\times$ & $\times$ & $\times$ \\
OpenCV       & \checkmark & \checkmark & $\times$ & $\times$ & $\times$ & $\times$ \\
Nvidia Dali & \checkmark & \checkmark & \checkmark & $\times$ & $\times$ & $\times$ \\
tf.image  & \checkmark & \checkmark & \checkmark & \checkmark & \checkmark & \checkmark \\
\lib{}   & \checkmark & \checkmark & \checkmark & \checkmark & \checkmark & \checkmark \\ \\
\end{tabular}
\caption{\label{tab:table-cv-frameworks} Comparison between  different computer vision libraries by their main features. \lib{} and tensorflow-image are the only frameworks that fully support batched operators and differentiable on the GPU.}
\end{table*}

In addition to introducing \lib, this paper also contributes with some demos showcasing how \lib{} and its components eases the implementation of several common computer vision tasks like image reconstruction, image registration, depth estimation or local features detection. This journal includes an extensive explanation of the different capabilities and algorithms that can be found in each of the different modules of the library, including short coding examples and links with fully working examples; we also include some additional experiments to benchmark against other frameworks and evaluate its usage depending on the batch size; and finally, we showcase an example about how easily can \lib{} be integrated within an end to end training system.

The rest of the paper is organized as follows: we review the state of the art in terms of open source software for computer vision and machine learning in Section \ref{section:related_work}; Section \ref{section:kornia} describes the design principles of the proposed library and all its components, and Section~\ref{section:use_cases} introduces use cases that can be implemented using the library's main features. \lib{} is public available in GitHub\footnote{\url{https://github.com/kornia/kornia}} with an Apache License 2.0 and can be installed in any Linux, MacOS or Windows operating system, having PyTorch as a single dependency, through the Python Package Index (PyPI) using the following command:

\mint[frame=single, framesep=5pt,  baselinestretch=1., style=vs, bgcolor=bg, fontfamily=courier, fontsize=\footnotesize]{bash}{pip install kornia}

\section{Related work}
\label{section:related_work}
In this section we  review  the state of the art for computer vision software libraries. Related works will be divided in two main categories: classical computer vision and deep learning oriented  frameworks. The former are focused on  the very first libraries that implement mostly algorithms optimized for the CPU, and the second category targets solutions for GPU.

\subsection{Classical computer vision libraries}
\label{section:related_work:traditional_vision}
Nowadays, there is a wide variety of options for frameworks that implement computer vision algorithms. However, during the early days of computer vision, it was difficult to find any available and free accessible software for image processing algorithms. All the existing software for computer vision was mostly developed within universities or at small teams in companies, and it was not shipped in any form nor released to the public domain. An example of this type of proprietary software specific for Machine Vision, was the Matrox Imaging Library (MIL)~\citep{matrox_imaging}.

It was not until Intel released the first version of the Open Source Computer Vision Library (OpenCV) that changed the paradigm for the existing Computer Vision software by that time making it accessible for everyone. OpenCV \citep{opencv} initially implemented computer vision algorithms for real-time ray tracing, visual interfaces and 3D display walls. All the algorithms were made available with a permissive library not only for research, but also for production and commercial usage. OpenCV changed the paradigm within the computer vision community given the fact that most   state of art algorithms in computer vision were now put in an common framework written very efficiently in C, becoming in that way a reference within the community.

The computer vision community shifted to improving or besting existing algorithms and started sharing their code with the community. This resulted in new code optimized mostly for CPU. Vedaldi et al. introduced \textit{VLFeat}~\citep{vedaldi08vlfeat}, an open source library that implements popular computer vision algorithms specializing in image understanding and local features extraction and matching. VLFeat was written in C for efficiency and compatibility, with interfaces in MATLAB. For ease of use, it supported Windows, Mac OS X, and Linux, and has been a reference for e.g. efficient implementations for local features extraction and matching such as Fisher Vector~\citep{Sanchez2013}, VLAD~\citep{VLAD2010}, SIFT~\citep{Lowe2004}, and MSER~\citep{Matas2002}.

MathWorks released a proprietary Computer Vision Toolbox inside one of its famous product MATLAB~\citep{MATLAB:2010} that covered many of the main computer vision, 3D vision, and video processing algorithms which has been used by many computer vision students and researchers becoming quite standard within the researcher community. The computer vision community have been using MATLAB for some decades, and many still use it.

Over time, new projects such as Numpy~\citep{oliphant2006guide} which includes powerful N-dimensional array objects manipulation and very optimised linear algebra support, started to position themselves as the angular stone for scientific computing within the Python language community. Another example is Scikit-learn \citep{scikit-learn} that with same philosophy as Numpy, partially implement machine learning algorithms  for classification, regression and clustering including support vector machines, random forests, gradient boosting and k-means. A similar project,  Scikit-image~\citep{scikit-image} implements open source collections of algorithms for image processing making it compatible with this new group of Python scientific packages.

\subsection{Deep learning and computer vision}
\label{section:related_work:deep_learning}
Computer vision frameworks have been optimized for CPU to fulfill realtime requirements for different applications, however, the recent success of deep learning changed the way how computer vision system are designed. A. Krizhevsky et al. introduced to the community \textit{Alexnet}~\citep{AlexNet2012} continuing the old ideas from Yann LeCun's Convolutional Neural Networks (CNNs)~\citep{MNIST1998} paper with an architecture similar to LeNet-5 and achieved the best results by far in The ImageNet Large Scale Visual Recognition Challenge (ILSVRC)  2012~\citep{ILSVRC15} image classification task. This was a breakthrough moment for the computer vision community, and changed the way computer vision was understood. In terms of software, new frameworks such \textit{Caffe} \citep{caffe}, Torch \citep{torch7}, MXNet~\citep{journals/corr/ChenLLLWWXXZZ15}, Chainer~\citep{Tokui:2019:CDL:3292500.3330756}, Theano~\citep{Bergstra10theano}, MatConvNet~\citep{Vedaldi15}, PyTorch~\citep{pytorch}, and Tensorflow \citep{tensorflow2015-whitepaper} appeared on the scene implementing more efficiently some of the classical operators for computer vision such as Convolutions in the GPU using parallel programming~\citep{CookCUDA} as an approach to handle the need for large amounts of data processing in order to train large deep learning models which in some cases overpass human performance e.g. in classifying objects.

With the rise of deep learning, most standard computer vision frameworks are now being used to perform pre-processing, or data augmentation on the CPU, which for some use cases like volumetric data in medical imaging or multi-spectral data is quite limited due to the need of parallelism to not decrease performance in both, training and inference time. Examples of libraries that are currently used to perform pre- and post-processing on the CPU within the deep learning frameworks are OpenCV or Pillow.

Given that most  deep learning frameworks still use standard vision libraries to perform the pre- and post-processing on CPU and similar to Tensorflow.image {as Table~\ref{tab:table-cv-frameworks} shows, \lib{} fill the gap} within the PyTorch ecosystem by introducing a computer vision library that implements standard computer vision algorithms taking advantage of the different properties that modern frameworks for deep learning like PyTorch can provide:

\begin{itemize}
    \setlength\itemsep{0.em}
	\item \textbf{differentiability} to avoid writing derivative functions for complex loss functions.
	\item \textbf{transparency} to perform parallel or serial computing in high-performance hardware devices in a common API.
	\item \textbf{distributed} to deploy large-scale applications in distributed systems.
	\item \textbf{production} ready to be used in efficient oriented deep learning applications.
\end{itemize}

\newpage

\section{\lib: Computer Vision for PyTorch}
\label{section:kornia}

\begin{figure*}[h]
\centering
\includegraphics[scale=0.30]{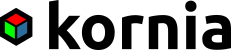}
\label{fig:kornia_logo}
\end{figure*}

\lib\footnote{\url{https://kornia.org}} can be defined as a computer vision library for PyTorch, inspired by OpenCV and with strong GPU support. \lib{} allows users to write code as if they were using native PyTorch providing high level interfaces to vision algorithms computed directly on tensors. In addition, some of the main PyTorch features are inherited by \lib{} such as a high performance environment with easy access to automatic differentiation, executing models on different devices (CPU, GPU or Tensor Porcessing Unit -- TPU), parallel programming by default, communication primitives for multiprocess parallelism across several computation nodes and code ready for production. In the following, we elaborate on these properties.

\noindent\textbf{Differentiable}. Any image processing algorithm that can be defined as a Direct Acyclic Graph (DAG) structure can be incorporated in a neural network and can be optimized during training, making use of the reverse-mode \citep{Speelpenning1980CFP} auto-differentiation \citep{Griewank:2008:EDP:1455489}, compute gradients via backpropagation \citep{kelley1960gradient}. In practice, this means that computer vision functions are operators that can be placed as layers within the neural networks for training via backpropagating through them.

\noindent\textbf{Transparent API}. A key component in the library design is its easy way to seamlessly add hardware acceleration to your program with a minimum effort. The library API is agnostic to the input source device, meaning that the algorithms can either be executed in several devices type such as CPU, GPU or the recently introduced TPU.

\noindent\textbf{Parallel programming}. Batch processing is another important feature that enables to run vision operators using data parallelism by default. The assumption for the operators is to receive N-channel image tensors as input batches, contrary to standard vision libraries with single 1-3 channel images. Hence, working with multispectral, hyperspectral or volumetric image can be done in a straight-forward manner is using \lib{}.

\noindent\textbf{Distributed}. Support for communication primitives for multi-process parallelism across several computation nodes running on one or more group of local or cloud based machines. The library design allows users to run their applications in different distributed systems, or even able to process large vision pipelines in an efficient way.

\noindent\textbf{Production}. Since version v1.0.0, PyTorch has the feature to serialize and optimize models for production purposes. Based on its just-in-time (JIT) compiler, PyTorch traces the models, creating \textit{TorchScript} programs at runtime in order to be run in a standalone C++ program using kernel fusion to do faster inference. This makes our library a perfect fit also for built-in vision products.

\subsection*{What \lib \hspace{1px} is NOT}

\lib{} aims to be a reimplementation for OpenCV for research purposes in the sense that mimics some of the main functionalities adding the ability to backpropagate through the different operators. However, note that Kornia does not seek to be a replacement for OpenCV since it is not optimised for production purposes or to achieve  high-performance in embedded devices. Even though the project is backed up by the OpenCV.org, there is no intention to merge in any form both projects in the mid-term. Kornia can be understood as a set of tools for training neural networks to be later used in production using other optimised frameworks.

\subsection{Library structure}
\label{section:kornia:library_structure}

The internal structure of the library is designed to cover different computer vision areas, including color conversions, low level image processing, geometric transformations and some utilities for training  such as specific  loss functions, conversions between data layouts for different frameworks, or functionalities to easily visualise images and debug models during   training. Similar to other frameworks, the library is composed of several sub-modules grouped by generic computer vision topics:

\newpage

\textbf{kornia.augmentations:} The library provides a fully functional set of routines that can be used to perform data augmentation for training deep networks. This module implements in a high level logic several functionalities found across the other modules. The main features of this  module, and similar to the rest of the library, is that can it perform data augmentation routines in a batch mode, using any supported device, and can be used for backpropagation. Some of the available functionalities which are worth to mention are the following: random rotations; affine and perspective transformations; several random color intensities transformations, image noise distortion, motion blurring, and many of the different differentiable data augmentation policies proposed in~\citep{DBLP:journals/corr/abs-2004-11966, huang20}. In addition, we include a novel feature which is not found in  other augmentations frameworks, which allows the user to retrieve the applied transformation or chained transformations after each call e.g. the generated random rotation matrix which can be used later to undo the image transformation itself, or to be applied to additional metadata such as the label images for semantic segmentation, in bounding boxes or landmark keypoints for object detection tasks. This is very valuable for research purposes since it gives  the user the flexibility to perform complex data augmentations pipelines. A snippet showcasing a small example is shown in Example 1. In addition, in section~\ref{section:use_cases:imgproc} we will review performance benchmarks for this module.

\vspace{.5cm}


%
\begin{tcolorbox}[every float=\centering, drop shadow, title=Example 1: Data augmentation pipeline, label={fig:examples:augmentation}]
    \includegraphics[width=1.\linewidth]{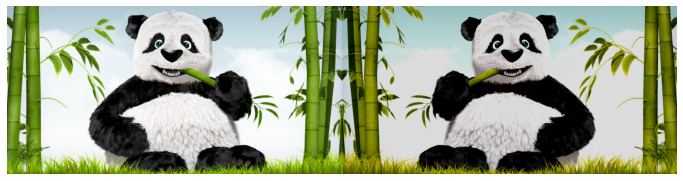}
    {Example showing how flexible is \lib{} to define a data augmentation pipeline using other PyTorch components. Concretely, this pipeline takes a batch of tensor images and randomly applies a vertical flip, retrieves the applied affine transformation, and finally applies a color jitter transformation based on the user defined preference. The code for this example can be found in the following \underline{\color{blue}\href{https://colab.research.google.com/drive/1iWUVw5jRBiUm0X48cROZO8yWCWN27sbG}{link}}.}
    \tcbsubtitle{Code}
\begin{minted}{python}
class DataAugmentationPipeline(torch.nn.Module):
    """Module to perform data augmentation using Kornia on tensors."""
    def __init__(self, apply_color_jitter: bool = True) -> None:
        super().__init__()
        self._apply_color_jitter = apply_color_jitter
        self.transforms = torch.nn.Sequential(
            K.augmentation.RandomVerticalFlip(
                p=0.5, return_transform=True
            ),
        )
        self.jitter = K.augmentation.ColorJitter(
            brightness=0.2, contrast=0.2, saturation=0.2, hue=0.1
        )

    @torch.no_grad()  # disable gradients for efficiency
    def forward(self, x: torch.Tensor):
        x_out, trans = self.transforms(x)
        if self._apply_color_jitter:
            x_out = self.jitter(x_out)
\end{minted}
\end{tcolorbox}

\newpage

\textbf{kornia.color:} Conversions between color spaces are useful when working with 3-band color images. For this purpose,  we provide several functionalities to perform operations that have a similar behaviour as those found in the existing libraries such as OpenCV~\citep{opencv} or Scikit-image~\citep{scikit-image}. We have introduced  small modifications in order to support floating point precision. The functionality found in this module covers operations to map back and forth between the most common color space representations, including Grayscale, RGB, RGBA, BGR, HSV, YCbCr, CIE-XYZ, CIE-Luv or CIE-Lab. In addition, we provide high level interfaces to manipulate color properties  to perform intensities normalisation, compute color histograms, or  adjust color properties like  the brightness, contrast, hue, gamma spectrum, saturation and blending operations to combine different images. We next show in Example 2, an example of how color conversion can be done.

\vspace{.5cm}

\begin{tcolorbox}[every float=\centering, drop shadow, title=Example 2: Color Space Conversion]
    \label{fig:examples:color}
    \includegraphics[width=1.\linewidth]{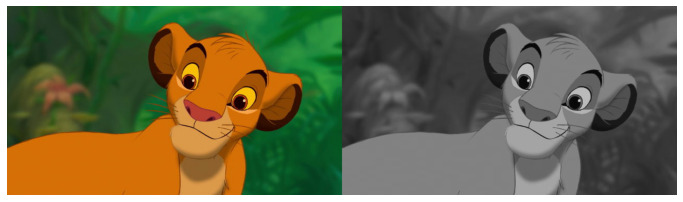}
    {Example showing how to load and decode an image using OpenCV and apply a color space conversion using \lib{}, and with a torch tensor image representation. The code for this example can be found in the following \underline{\color{blue}\href{https://colab.research.google.com/drive/1dgT9-QLZiTjxPK9Ej4YnyV953uTQKBCm}{link}}.}
    \tcbsubtitle{Code}
\begin{minted}{python}
# load image in numpy using OpenCV
img: np.ndarray = cv2.imread("simba.png", cv2.IMREAD_COLOR)

# load image in torch.Tensor
img_bgr: torch.Tensor = K.image_to_tensor(img)
img_rgb = K.bgr_to_rgb(img_bgr)
img_rgb = K.normalize(img_rgb, 0., 255.)

# apply color transforms
img_gray = K.rgb_to_grayscale(img_rgb)
\end{minted}
\end{tcolorbox}

\newpage

\textbf{kornia.features:} Local features detection and description are a key ingredient in a wide range of computer vision algorithms, for e.g. image stitching, structure from motion, or image retrieval. \lib{} provides operators to detect local features, compute descriptors, and perform feature matching. The module contains differentiable versions of the Harris corner detector~\citep{Harris88}, Shi-Tomasi corner detector detector~\citep{Shi94goodfeatures}, Hessian detector~\citep{Hessian78},  their scale and affine covariant versions~\citep{Mikolajczyk2004}, DoG~\citep{Lowe2004}, patch dominant gradient orientation~\citep{Lowe2004} and the SIFT descriptor~\citep{Lowe2004} and recent deep learning based methods such as HardNet~\citep{mishchuk2017working} or SOSNet~\citep{tian2019sosnet}. In addition, this module provides a high level API to perform detections in scale-space, where classical hard non-maxima suppression is replaced with its soft version, similar to the recently proposed Multiscale Index Proposal layer (M-SIP)~\citep{KeyNet2019}, one can seamlessly replace any or all modules with deep learned counterparts. We also provide a set of utility functions for working with different local features geometry representations and conversion between them: oriented and non-oriented circles, local affine frames (LAFs) and oriented ellipses as described in~\citep{Mikolajczyk2004}. The use of this functionality is shown in Example 3.

\vspace{.5cm}

\begin{tcolorbox}[every float=\centering, drop shadow, title=Example 3: Feature detection and Matching]
    \label{fig:examples:feature}
    \includegraphics[width=1.\linewidth]{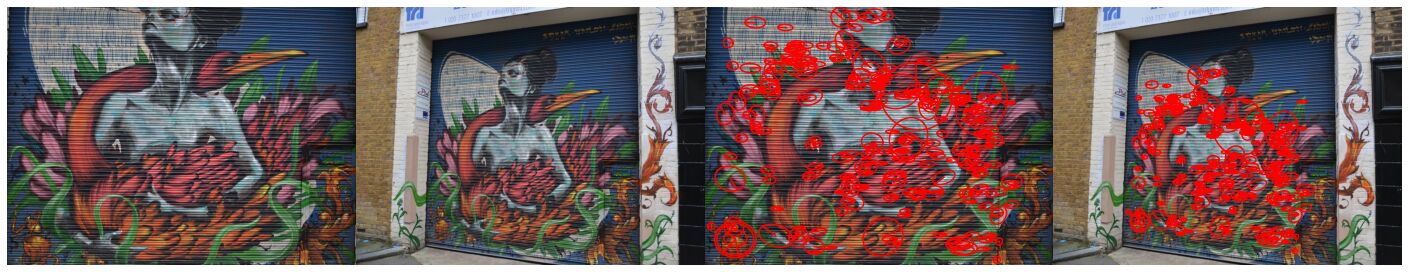}
    {Example showing the use of several components for local feature detection, matching and geometry verification. In this case, we apply a standard Hessian blob detector in the Space scale including Non-maxima suppression, mutual HardNet~\citep{mishchuk2017working} features matching plus a step for estimating the homography using DLT algorithm. The code for this example can be found in the following \underline{\color{blue}\href{https://colab.research.google.com/drive/1hl6ex1i8eNaoIZICxNqIO40PGcb8Txdo}{link}}.}
    \tcbsubtitle{Code}
\begin{minted}{python}
# Define local deature detector and descriptor
PS = 32  # patch size
detector = K.ScaleSpaceDetector(
    num_features=2000,
    resp_module=K.BlobHessian(),
    nms_module=K.ConvQuadInterp3d(10),
    scale_pyr_module=K.ScalePyramid(3, 1.6, PS, double_image=False),
    ori_module=K.LAFOrienter(19),
    aff_module=K.LAFAffineShapeEstimator(19),
    mr_size=6.0)

descriptor = K.HardNet(pretrained=True)

# detect and extract patches
lafs, resps = detector(timg_gray)
patches =  K.extract_patches_from_pyramid(timg_gray, lafs, 32)
B, N, CH, H, W = patches.size()
descs = descriptor(patches.view(B * N, CH, H, W)).view(B, N, -1)

# Matching
tentatives, scores = K.match_smnn(descs[0], descs[1], 0.95)
kps = KF.laf.get_laf_center(lafs)
kps_tent1 = kps[0:1,tentatives[:,0]]
kps_tent2 = kps[1:2,tentatives[:,1]]

# Finding homography
H = K.find_homography_dlt_iterated(
    kps_tent1, kps_tent2, 1-scores.view(1,-1)
)
\end{minted}
\end{tcolorbox}

\newpage

\textbf{kornia.filters:} Image filtering is another traditional operation  in image processing and computer vision used in a number applications, from  noise removal to fancy work-arts creation. This module provides operators to perform linear and non-linear filtering operations on tensor images. High level functions to convolve tensors with hand-crafted kernels; for computing first and second order image or n-dimensional tensor derivatives; high level differentiable implementations for blurring algorithms such as Gaussian, Box, Median or Motion blurs; Laplace, and Sobel\citep{kanopoulos1988design} edges detector. The functionalities found in this module can be either used for creating accelerated computer pipelines or, as we will see in section~\ref{section:use_cases:image_reconstruction}, to compute loss functions to maintain image properties during reconstructions processes. Example 4 shows a use-case of this functionality.

\vspace{.5cm}

\begin{tcolorbox}[every float=\centering, drop shadow, title=Example 4: Image filtering and Edge detection]
    \label{fig:examples:filters}
    \includegraphics[width=1.\linewidth]{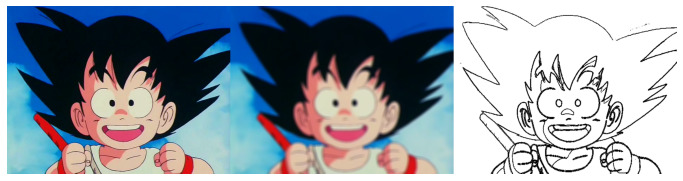}
    {Example about how to load an image using OpenCV, apply  2D Gaussian blur filtering and computing the Sobel edges on a RGB torch tensor image. The code for this example can be found in the following \underline{\color{blue}\href{https://colab.research.google.com/drive/1IiNHo5TjgQShrI7XoMOtam3PyZ9FNToh}{link}}.}
    \tcbsubtitle{Code}
\begin{minted}{python}
# load image in numpy using OpenCV
img: np.ndarray = cv2.imread("goku.png", cv2.IMREAD_COLOR)

# load image in torch.Tensor
img_bgr: torch.Tensor = K.image_to_tensor(img, keepdim=False)
img_rgb = K.bgr_to_rgb(img_bgr).float() / 255.
img_gray = K.rgb_to_grayscale(img_rgb)

# apply a gaussian blur
img_edge = K.sobel(img_gray)
img_blur = K.gaussian_blur2d(img_rgb, (11, 11), (10.5, 10.5))
\end{minted}
\end{tcolorbox}

\newpage

\textbf{kornia.geometry:} Geometric image transformations is another key ingredient in computer vision to manipulate images. Since geometry operations are typically performed in 2D or 3D,  we provide several algorithms to work with both cases. This module, the original core of the library, consists of the following submodules: \textbf{transforms}, \textbf{camera}, \textbf{conversions}, \textbf{linalg} and \textbf{depth}. We next describe each of them.

\begin{itemize}
    \item \textbf{\mintinline{python}|transforms:|} The module provides low level interfaces to manipulate 2D images,  with routines for Rotating, Scaling, Translating, Shearing; Cropping functions in several modalities such as central crops, crop and resize; Flipping transformations in the vertical and horizontal axis; Resizing operations; Functions to warp tensors given affine or perspective transformations, and utilities to compute the transformation matrices to perform the mentioned operations (see Example 5 below).
    \item \textbf{\mintinline{python}|camera:|} A set of routines specific to different types of camera representations such as Pinhole or Orthographic models containing functionalities such as projecting and unprojecting points from the camera to a world frame.
    \item \textbf{\mintinline{python}|conversions:|} Routines to perform conversions between angle representation such as radians to degrees, coordinates normalization, and homogeneous to euclidean. Moreover, we include advanced conversions for 3D geometry representations such as Quaternion, Axis-Angle, RotationMatrix, or Rodrigues formula.
    \item \textbf{\mintinline{python}|linalg:|} Functions to perform general rigid-body homogeneous transformations. We include implementations to transform points between frames and for homogeneous transformations, manipulation such as composition, inverse and to compute relative poses.
    \item \textbf{\mintinline{python}|depth:|} A set of layers to manipulate depth maps such as how to compute 3D point clouds given depth maps and calibrated cameras; compute  surface normals per pixel and warp tensor frames given calibrated cameras setup.
\end{itemize}

\begin{tcolorbox}[every float=\centering, drop shadow, title=Example 5: Geometric Image transformation]
    \label{fig:examples:geometry}
    \includegraphics[width=1.\linewidth]{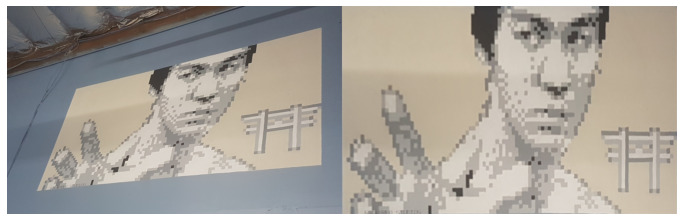}
    {Example showing how to compute the perspective transformation matrix to warp one image into another given four control points. The code for this example can be found in the following \underline{\color{blue}\href{https://colab.research.google.com/drive/1TgRiOs9x0W98axsb9jDTnYul0pGCeGQv}{link}}.}
    \tcbsubtitle{Code}
\begin{minted}{python}
# the source points are the region to crop corners
points_src = torch.tensor([[
    [125., 150.], [562., 40.], [562., 282.], [54., 328.],
]])

# the destination points are the image vertexes
h, w = img_rgb.shape[-2:]  # destination size
points_dst = torch.tensor([[
    [0., 0.], [w - 1., 0.], [w - 1., h - 1.], [0., h - 1.],
]])

# compute perspective transform
M: torch.tensor = K.get_perspective_transform(points_src, points_dst)

# warp the original image by the found transform
img_warp: torch.tensor = K.warp_perspective(img_rgb, M, dsize=(h, w))
\end{minted}
\end{tcolorbox}

\textbf{kornia.losses:} This submodule contains a compilation of specific losses for computer vision to solve problems in different areas like semantic segmentation including the \textit{FocalLoss} ~\citep{DBLP:journals/corr/abs-1708-02002}, \textit{DiceLoss} or \textit{TverskyLoss}; for image synthesis we include metrics that can also be used for optimisation such as the Structural Similar Index Loss (SSIM)~\citep{Wang:2004:IQA:2319031.2320551}, the Peak signal-to-noise ratio (PSNR) or the Total Variation typicaly used in  image denoising. We also include interfaces that can be used for optimizing over 1d or 2d distributions with the \textit{Kullback-Leiber} or \textit{Jensen-Shannon} divergence losses. See Example 6.

\vspace{.25cm}

\begin{tcolorbox}[every float=\centering, drop shadow, title=Example 6: Computer vision losses]
    \label{fig:examples:losses_snipet}
    Pseudo-code example showing the usage of some of the different loss functions available in the library. In this case, given two tensor images we show how to compute the losses typically used in image reconstruction synthesis and semantic segmentation  tasks.
\begin{minted}{python}
# segmentation
loss = kornia.focal_loss(predictions, labels)
loss = kornia.dice_loss(predictions, labels)
loss = kornia.tversky_loss(predictions, labels)

# reconstruction
loss = kornia.psnr_loss(img1, img2)
loss = kornia.ssim_loss(img1, img2, window_size=5)
\end{minted}
\end{tcolorbox}

\vspace{.25cm}

\textbf{kornia.contrib:} Inspired by other libraries, we also include a  contrib module  to collect experimental operators and user contributions containing routines for splitting tensors in blocks (see Example 7), or recent paper implementations such as modules combining learned features with gaussian blurring that can be inserted within the networks to improve stability and robustness against image shifting ~\citep{zhang2019shiftinvar}.

\vspace{.25cm}

\begin{tcolorbox}[every float=\centering, drop shadow, title=Example 7: Extract image patches]
    \label{fig:examples:contrib}
    \includegraphics[width=1.\linewidth]{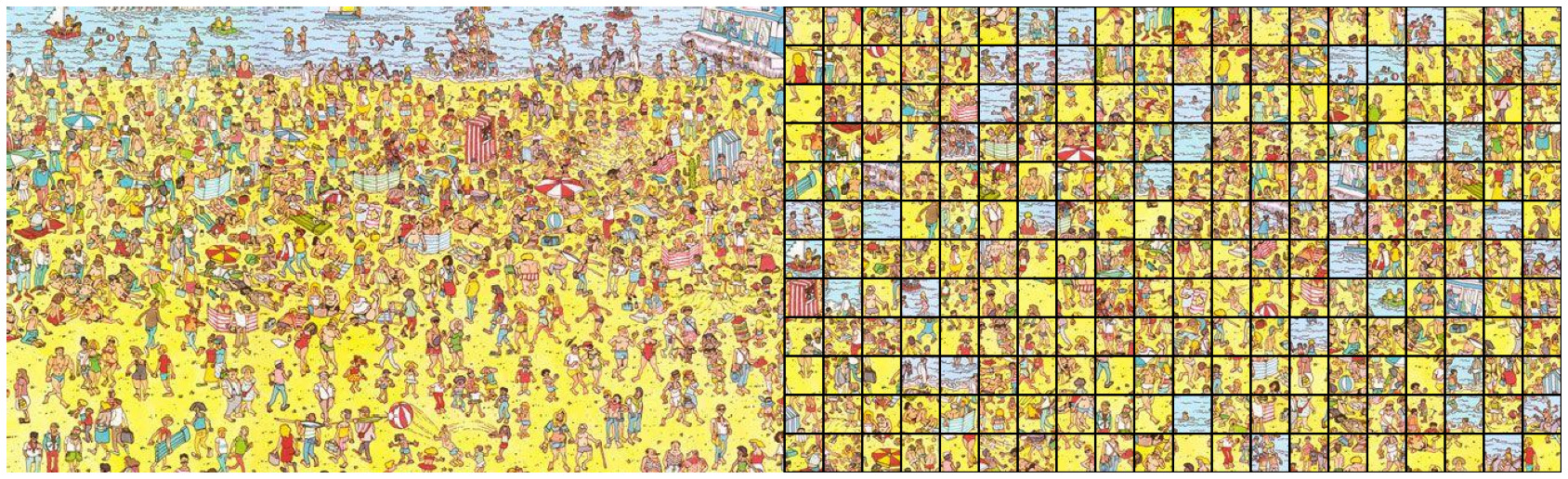}
    {Example showing how to split an image tensor into patches, sort the patches in order to create a mosaic. The code for this example can be found in the following \underline{\color{blue}\href{https://colab.research.google.com/drive/1JEPVShMILiFd4VfbJ5P3qTrqajSfR0Xv}{link}}.}
    \tcbsubtitle{Code}
\begin{minted}{python}
# load image in numpy using OpenCV
img: np.ndarray = cv2.imread("wally.jpg", cv2.IMREAD_COLOR)

# load image in torch.Tensor
img_bgr: torch.Tensor = K.image_to_tensor(img, keepdim=False)
img_rgb = K.bgr_to_rgb(img_bgr)
img_rgb = K.normalize(img_rgb.float(), 0., 255.)

# extract tensor patches
patches = K.extract_tensor_patches(
    img_rgb, window_size=32, stride=32
)
\end{minted}
\end{tcolorbox}

\section{Use cases}
\label{section:use_cases}
This section presents practical examples of the library use for well known classical vision problems demonstrating its easiness for computing the derivatives of complex loss functions and releasing the user of that part. We first show quantitative and qualitative results on experiments comparing our image processing API compared to other existing image processing libraries. We then provide an end to end training example for low-dimensional embedding application showcasing the usage of image processing functions as loss functions. Next, we describe an example of image registration
leveraging on  our differentiable warpers in a multi-scale fashion. Finally, we provide an example demonstrating the applicability of  our differentiable local features implementations to solve a classical wide baseline stereo matching problem.

\subsection{Batched image processing}
\label{section:use_cases:imgproc}

In Section \ref{section:related_work} we reviewed existing libraries implementing classical image processing algorithms optimized for practical applications such noise reduction, image enhancement and restoration. We next  want to use our framework for similar purposes. In addition, we include benchmarks against other existing vision libraries showing that \lib{} achieved competitive performances in terms of the time performance.

As stated in section \ref{section:kornia:library_structure}, \lib{} provides implementations for low level image processing, e.g. color conversions, filtering and geometric   transformations that implicitly use native PyTorch operators such as 2D convolutions and simple matrix multiplications,  all  optimized for different hardware devices. Qualitative results of our image processing API are illustrated in Figure~\ref{fig:imgproc}. Our API can be combined with other PyTorch components allowing to run vision algorithms via parallel programming, or even sending composed functions to distributed environments.

\begin{figure*}
\centering
 \begin{minipage}{.51\textwidth}
    \includegraphics[scale=0.41]{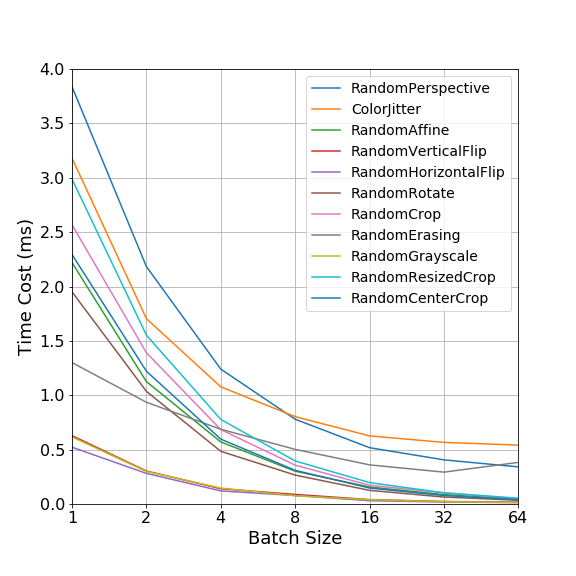}
 \end{minipage}%
\begin{minipage}{.49\textwidth}
    \centering
    \includegraphics[scale=0.63]{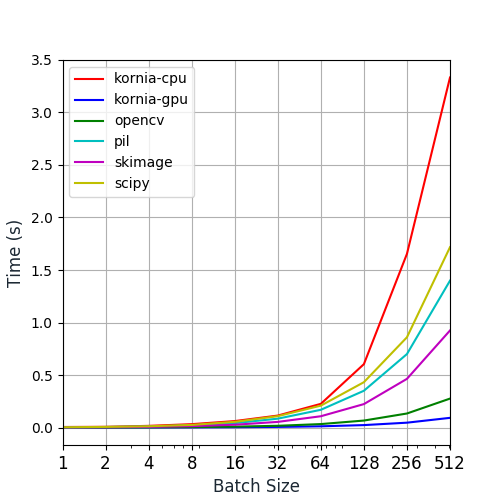}
\end{minipage}
\caption{{\bf Computation time comparison of Kornia with other libraries. }\textbf{Left:} Operation-wise per-sample timing benchmark with different batch sizes with fixed image size 224x224. Obviously, with the increasing batch sizes, the per-sample augmentation speed is boosted dramatically. Also, the per-sample augmentation speed is gradually stabilized after batch size is increased above 16. In this sense, \lib{} can better accelerate the data augmentation process when using batch sizes larger than 16. \textbf{Right:} Results of the benchmark comparing \lib{} to other state-of-the-art vision libraries. We measure the elapsed time for computing Sobel edges (lower is better).}
\label{fig:op_benchmark}
\end{figure*}

\textbf{Benchmark.} The scope of this library is not to provide explicit optimized code for computer vision, but we want to show an experiment comparing the performance of our library with respect to other existing vision libraries, namely OpenCV \citep{opencv}, PIL, skimage \citep{scikit-image} and scipy \citep{scikit-learn}, see Figure \ref{fig:op_benchmark}. The purpose of this experiment is to give a brief idea of how our implementations compare to libraries that are very well optimized for specific computer vision algorithms. The setup of the experiment assumes as input an RGB tensor of images with a fixed resolution of (256x256), and varying the size of the batch. In this experiment, we compute Sobel edges 500 times measuring the median elapsed time between samples. The results show that for small batches, \lib's performance is similar to those obtained using  other libraries. It is worth noting that when we use a large batch size, the performance for our CPU implementation is the lowest, but when using the GPU we get the best timing performance. The machine used for this experiment was an Intel(R) Xeon(R) CPU E5-1620 v3 @ 3.50GHz and a Nvidia Geforce GTX 1080 Ti.

\textbf{Operation-wise benchmark.} The second experiment evaluated the  Kornia augmentation module, which is designed and optimized for batch accelerated augmentations. We firstly assessed the operation-level time-efficiency of Kornia and TorchVision with batch size 1, under   image sizes ranging from 224x224 to 600x600. As shown in Table~\ref{tab: op_kornia_vs_torchvision}, the time perfomance of Kornia and TorchVision are comparable when process single RGB image. Next, we further tested the Kornia augmentation module in batch mode. Figure~\ref{fig:op_benchmark} shows the results of the per-sample time cost for each operation. With  larger batch sizes, the per-sample augmentation speed is boosted dramatically. In light of these  results, it is recommended to have batch sizes larger than 16 to make best use of \lib{}. The performances were measured using Intel Xeon E5-2698 v4 2.2 GHz (20-Core) and a Nvidia Tesla V100.

\begin{table}
\centering
\small
\begin{tabular}{llrrrrrrrr}

\toprule
     &     Image Size        &     224 &     240 &     260 &     300 &     380 &     456 &     528 &     600 \\
\cmidrule{1-2}
Operation & Library &        &        &        &        &        &        &        &        \\
\midrule
ColorJitter & Kornia &  12.73 &  \bfseries11.33 &  \bfseries12.13 &  13.30 &  \bfseries15.13 &  17.46 &  \bfseries19.87 &  \bfseries22.94 \\
                   & TorchVision &  \bfseries11.36 &  12.04 &  12.64 &  13.30 &  15.31 &  \bfseries17.30 &  20.18 &  23.03 \\
\midrule
RandomAffine & Kornia &   \bfseries2.22 &   \bfseries2.14 &   \bfseries2.22 &   \bfseries2.35 &   \bfseries2.27 &   \bfseries2.21 &   2.32 &   \bfseries2.19 \\
                   & TorchVision &   2.74 &   2.51 &   2.61 &   2.55 &   2.35 &   2.22 &   \bfseries2.22 &   2.26 \\
\midrule
RandomCenterCrop & Kornia &   \bfseries2.29 &  \bfseries 2.26 &   \bfseries2.25 &   \bfseries2.38 &   \bfseries2.31 &   2.40 &   \bfseries2.31 &   2.40 \\
                   & TorchVision &   2.48 &   2.48 &   2.37 &   2.49 &   2.44 &   \bfseries2.38 &   2.45 &   \bfseries2.34 \\
\midrule
RandomCrop & Kornia &   \bfseries2.57 &   \bfseries2.57 &   \bfseries2.57 &   2.72 &   \bfseries2.59 &   2.64 &   \bfseries2.52 &   2.56 \\
                   & TorchVision &   2.90 &   2.72 &   2.59 &   \bfseries2.65 &   2.60 &   \bfseries2.59 &   2.65 &   2.56 \\
\midrule
RandomErasing & Kornia &   \bfseries1.30 &   \bfseries1.33 &   1.46 &   \bfseries1.48 &  \bfseries 1.65 &   1.99 &   2.36 &   2.68 \\
                   & TorchVision &   1.51 &   1.35 &   \bfseries1.41 &   1.52 &   1.93 &   \bfseries1.94 &   \bfseries2.25 &   \bfseries2.58 \\
\midrule
RandomGrayscale & Kornia &   \bfseries0.62 &   \bfseries0.56 &   \bfseries0.60 &   \bfseries0.56 &   0.59 &   \bfseries0.57 &   0.74 &   \bfseries0.57 \\
                   & TorchVision &   0.68 &   0.59 &   0.61 &   0.59 &   0.59 &   0.59 &   \bfseries0.72 &   0.60 \\
\midrule
RandomHorizontalFlip & Kornia &   \bfseries0.52 &   \bfseries0.54 &   0.52 &   0.58 &   \bfseries0.47 &   \bfseries0.48 &   0.64 &   \bfseries0.48 \\
                   & TorchVision &   0.56 &   0.58 &   \bfseries0.47 &   \bfseries0.48 &   0.49 &   0.49 &   \bfseries0.48 &   0.51 \\
\midrule
RandomPerspective & Kornia &   \bfseries3.84 &   4.02 &   \bfseries4.02 &   4.27 &   \bfseries4.34 &   \bfseries5.05 &   5.55 &   6.25 \\
                   & TorchVision &   4.70 &   \bfseries3.88 &   4.08 &   \bfseries3.99 &   4.44 &   5.14 &   \bfseries5.44 &   \bfseries5.73 \\
\midrule
RandomResizedCrop & Kornia &   \bfseries2.99 &   \bfseries2.92 &   \bfseries2.98 &   \bfseries2.88 &   3.05 &   \bfseries3.02 &   \bfseries2.89 &   \bfseries2.88 \\
                   & TorchVision &   3.27 &   2.97 &   3.15 &   2.96 &   \bfseries3.04 &   3.04 &   2.96 &   2.97 \\
\midrule
RandomRotate & Kornia &   \bfseries1.95 &   \bfseries1.84 &   \bfseries1.87 &   2.01 &   \bfseries1.95 &   \bfseries1.90 &   \bfseries1.94 &   2.02 \\
                   & TorchVision &   2.08 &   1.93 &   2.08 &   \bfseries1.92 &   2.01 &   1.94 &   2.07 &   \bfseries1.93 \\
\midrule
RandomVerticalFlip & Kornia &   \bfseries0.63 &   0.71 &   0.62 &  \bfseries 0.55 &   \bfseries0.59 &   0.57 &   \bfseries0.62 &   0.65 \\
                   & TorchVision &   0.70 &   \bfseries0.62 &   \bfseries0.55 &   0.56 &   0.60 &   \bfseries0.56 &   0.65 &   \bfseries0.61 \\
\bottomrule
\end{tabular}
\vspace{.5cm}
\caption{\label{tab: op_kornia_vs_torchvision} {\bf Performance time comparison of Kornia and TorchVision using different image sizes.} The results are computed as the average time cost (milliseconds) of 10 runs for each operation. Note that  both libraries have very similar computation time performances. 
}
\end{table}

\subsection{End to end Low-dimensional embedding}
\label{section:use_cases:image_reconstruction}

Encoding images into low-dimensional spaces is another well know topic in computer vision that has been studied for many years. The current trend  addresses the problem using  Convolutional Autoencoder architectures and  its   variants, such as Variational Auto Encoders (VAE). However, many of the proposed methods do not take into account image properties in the decoding phase. In this example, we  showcase how easily \lib{} components could allow to introduce these properties while training the networks, and exploit the differentiability property to enforce e.g. robustness in the edges reconstruction, or even make use of the color properties to go from one color space to another and backpropagate the gradients using those constraints.

\begin{figure*}
    \begin{center}
        \begin{tabular}{c}
        \includegraphics[width=\textwidth]{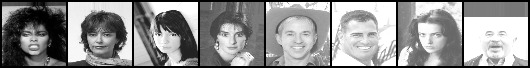} \\
        \includegraphics[width=\textwidth]{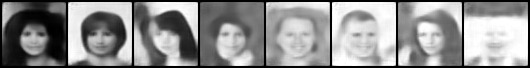} \\
        \includegraphics[width=\textwidth]{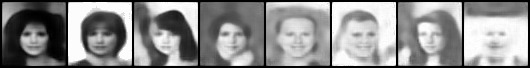} \\
        \end{tabular}
    \end{center}
    \caption{Results obtained in the experiment for learning low-dimensional embeddings and using them to decode the original image using the equation loss (\ref{eq:synthesis_loss1}). \textit{Row 1:} The original images used as input for the network. \textit{Row 2:} the reconstructed images using $\alpha = 1$ which means that the network optimises for the L1 distance. \textit{Row 3:} the reconstructed images using $\alpha = 0.5$ where the network optimises for the L1 distance at the same time as for the Sobel Edges. We can be appreciate that the images in the last row keep the structure of the edges in the central part of the faces.}
    \label{fig:synthesis}
\end{figure*}

\textbf{Implementation.} In order to showcase the explained problem, we have taken the architecture from one of the state of the art methods for image encoding \textit{Deep Feature Consistent Variational Autoencoder} ~\citep{DBLP:journals/corr/HouSSQ16a} slightly modifiying the network to get rid of the variational part. For training, we have chosen a popular dataset for this task - \textit{CelebA} dataset ~\citep{liu2015faceattributes} which is made of a set of images with aligned and cropped faces of famous celebrities. The experiment consists in evaluating the network qualitatively in a validation set composed by 30\% of the images from the original set. Our baseline to compare with consists of a loss function that optimises the L1 distance between the input image and the reconstructed by the network in RGB color space. We further evaluate the performance by adding constraints to the network during the learning process. Concretely, in this toy example we propose to add extra losses that aim  to keep the Sobel edges between the original and the reconstructed image. As we can see in Figure~\ref{fig:synthesis}, when the network is trained using a single loss, the reconstructed images look more smooth respect to when we train using a composed loss function including the Sobel edges detector as a variable to optimize during the reconstruction. The final loss we used has the following shape:

\begin{equation}
    \label{eq:synthesis_loss1}
	Loss = \alpha {\mid I - \hat{I} \mid} + (1 - \alpha) * {\mid \text{Sobel}(I) - \text{Sobel}(\hat{I}) \mid}
\end{equation}

This example opens a full research path to explore all these possibilities towards using very known image properties constraints within our neural network guiding the gradients towards a more realistic solution. The intention of this experiment is not to provide a solution to the problem since it might require a complex grid search parameters search, but instead,  to illustrate researchers how to build sophisticated loss functions based on \lib{} differentiable components for image processing.

\subsection{Image registration by Gradient Descent}
\label{section:use_cases:homography_estimation}
In the following, we show the potential of the library for tasks reasoning about the 2D planar geometry (e.g. marker-based camera pose estimation or spatial transformer networks~\citep{jaderberg2015spatial}). \lib{} provides a set of differentiable operators to perform geometric image transformations such as rotations, translations, scalings, shearings, as well as affine and homography transformation. At the core of the geometry module, we have implemented an operator \mintinline{python}|kornia.HomographyWarper|, which, based on the homography matric, warps a tensor in the reference frame A to a reference frame B  in a very efficient way.

\textbf{Implementation.} The task to solve is image registration using a multi-scale version of the Lucas-Kanade~\citep{BaM2004} strategy. Given a pair of images $I_a$ and $I_b$, it optimizes the parameters of the homography $H_a^b$ that minimizes the photometric error between $I_b$ and the transformation of $\hat{I_b}$ denoted as $\omega(I_a, H_a^b)$. Thanks to the Pytorch \textit{Autograd} engine, this can be implemented without explicitly computing the derivatives of the loss function from equation \ref{eq:homography_loss}, resulting in a very compact and intuitive code.
\begin{equation}
\text{Loss} = \sum_{u,v}^{N} \| I_b - \omega(I_a, H_a^b) \|_1
\label{eq:homography_loss}
\end{equation}

\begin{figure*}
    \setlength\tabcolsep{2.5pt}
    \begin{center}
        \begin{tabular}{c c c c c c}
        \textbf{Level 1} & \textbf{Level 2} & \textbf{Level 3} & \textbf{Level 4} & \textbf{Level 5} & \textbf{Level 6} \\
        \includegraphics[width=2.5cm]{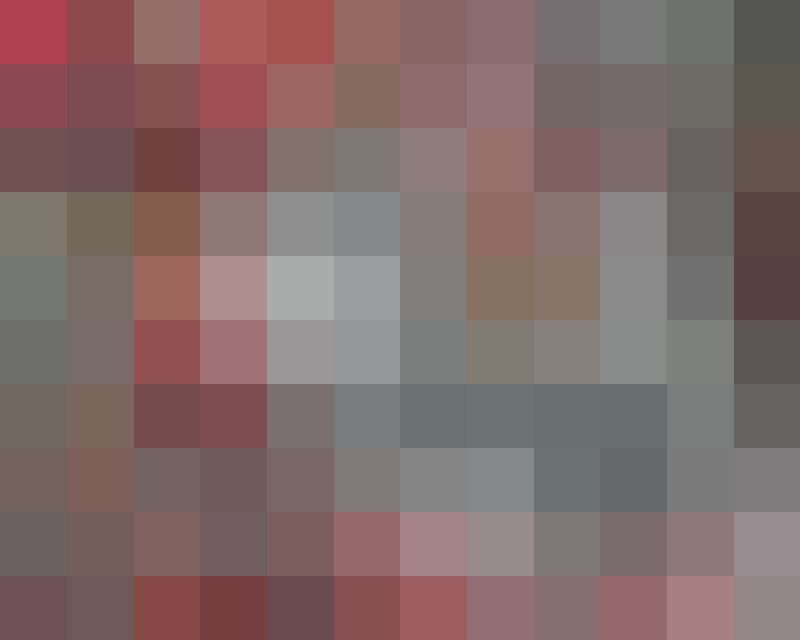} &
        \includegraphics[width=2.5cm]{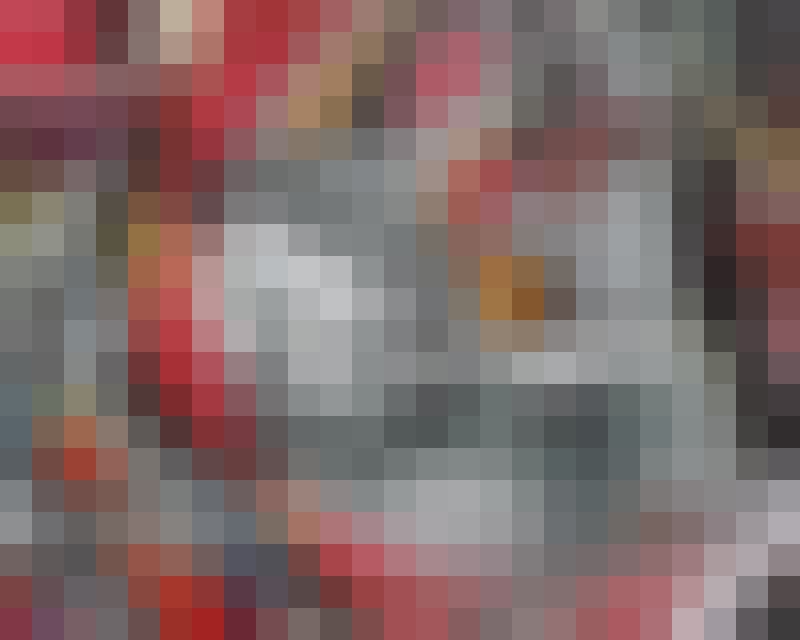} &
        \includegraphics[width=2.5cm]{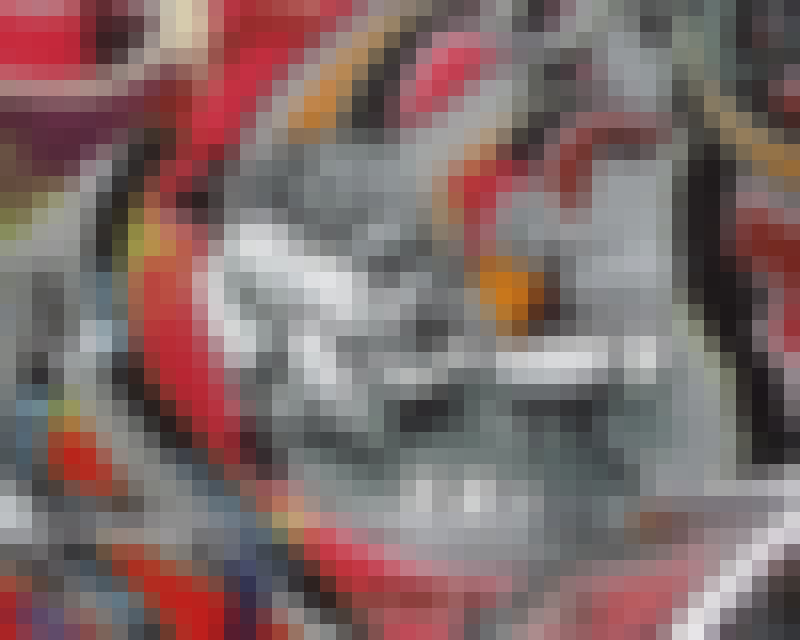} &
        \includegraphics[width=2.5cm]{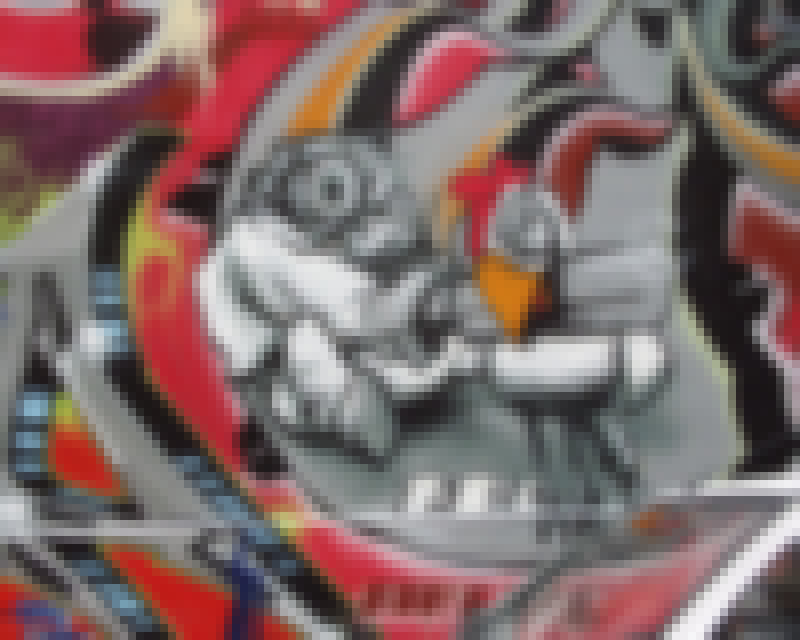} &
        \includegraphics[width=2.5cm]{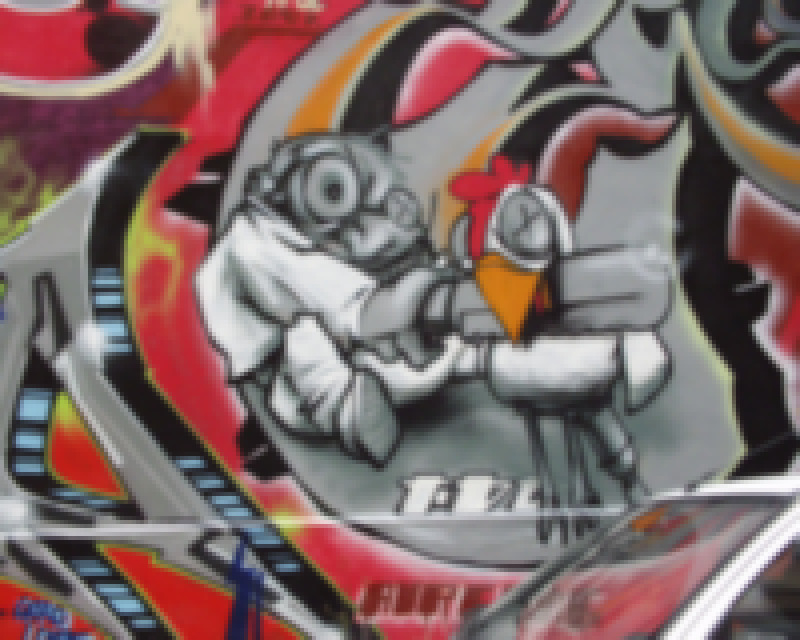} &
        \includegraphics[width=2.5cm]{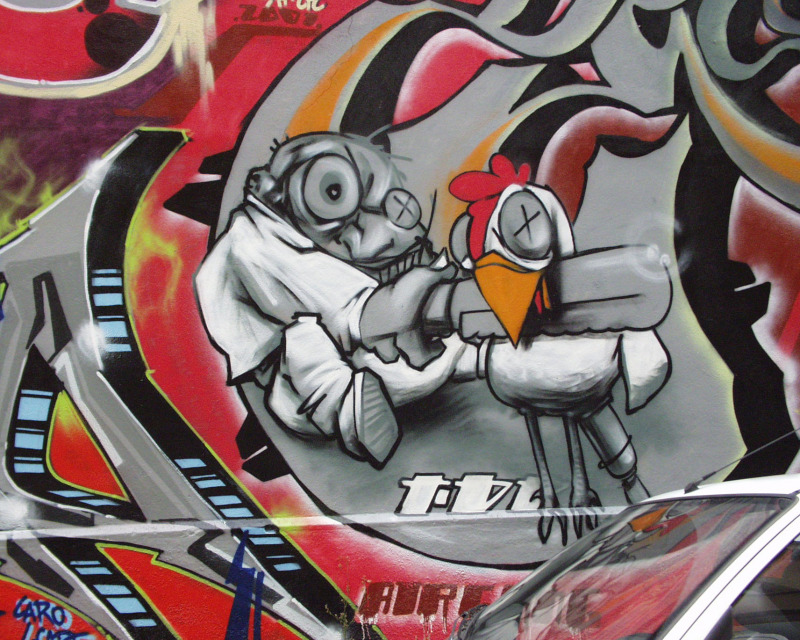} \\
        \includegraphics[width=2.5cm]{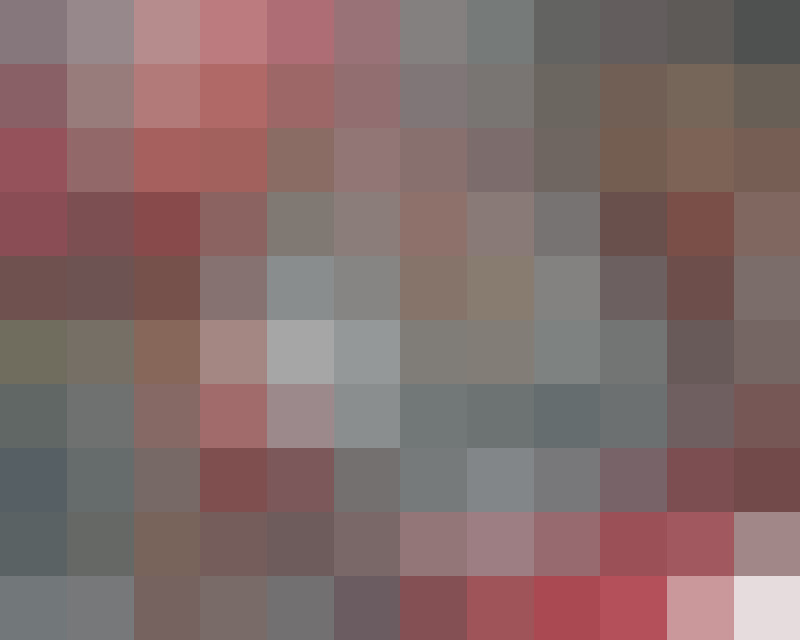} &
        \includegraphics[width=2.5cm]{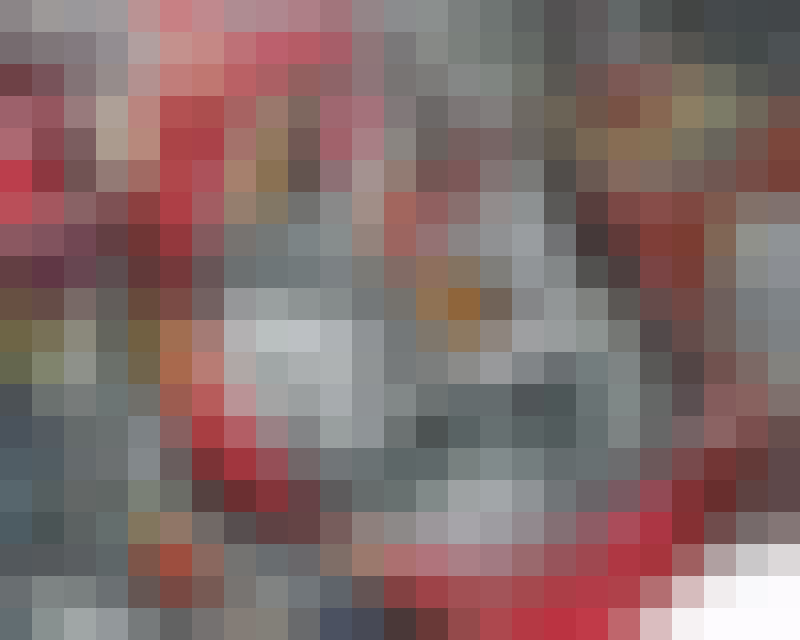} &
        \includegraphics[width=2.5cm]{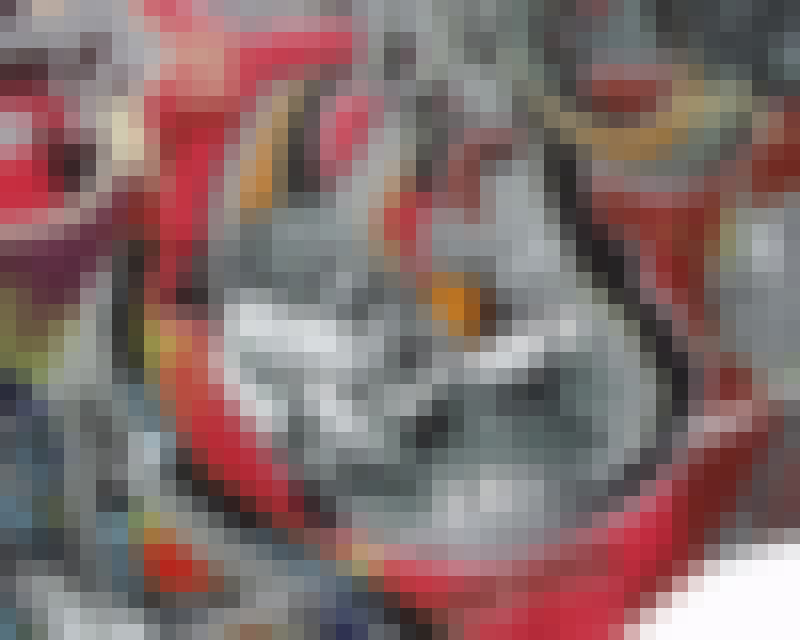} &
        \includegraphics[width=2.5cm]{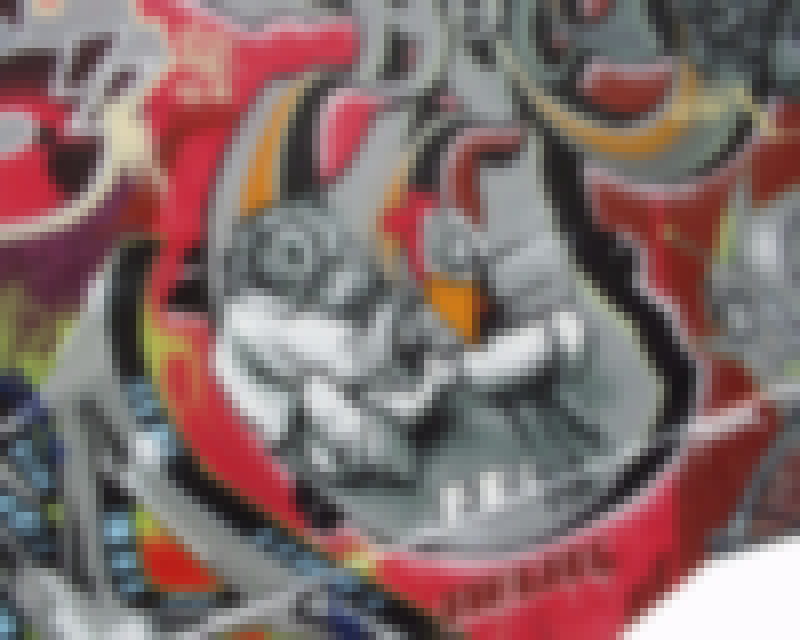} &
        \includegraphics[width=2.5cm]{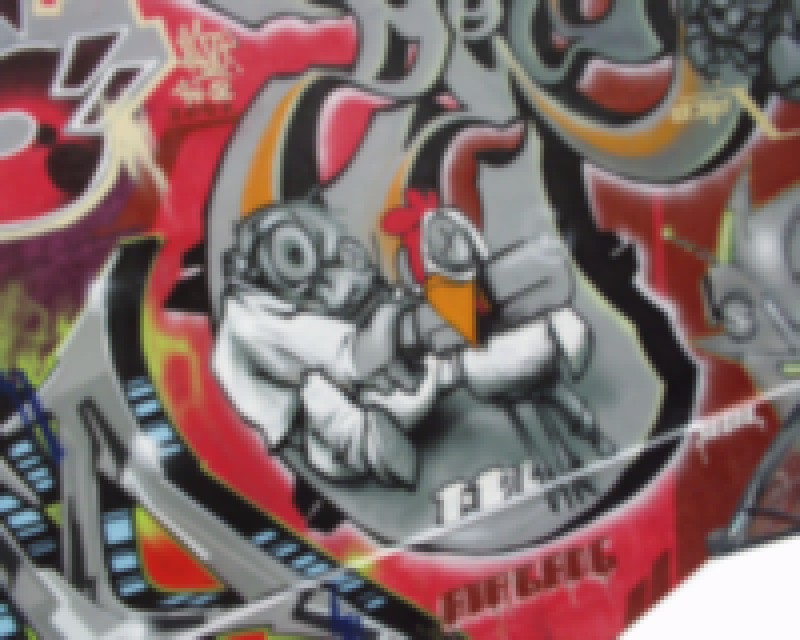} &
        \includegraphics[width=2.5cm]{data/homography/img_dst_7.jpg} \\
        \includegraphics[width=2.5cm]{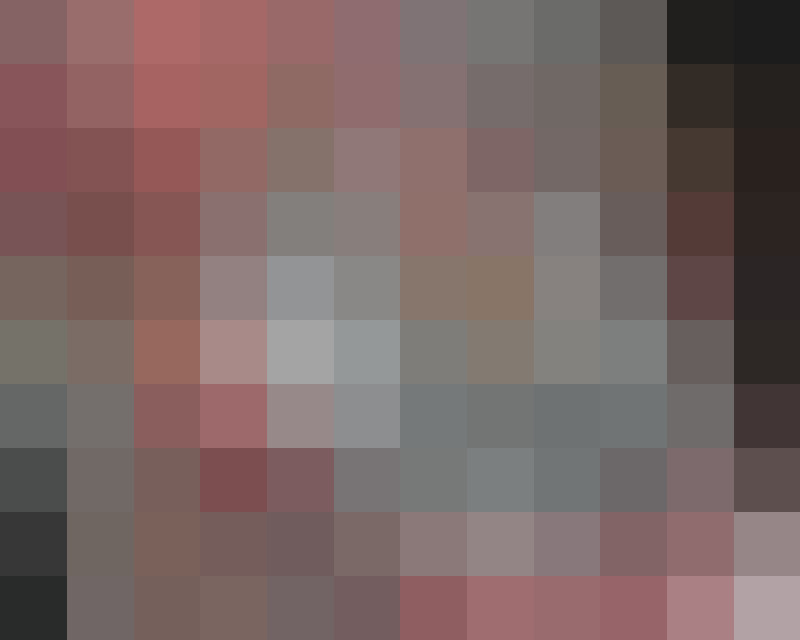} &
        \includegraphics[width=2.5cm]{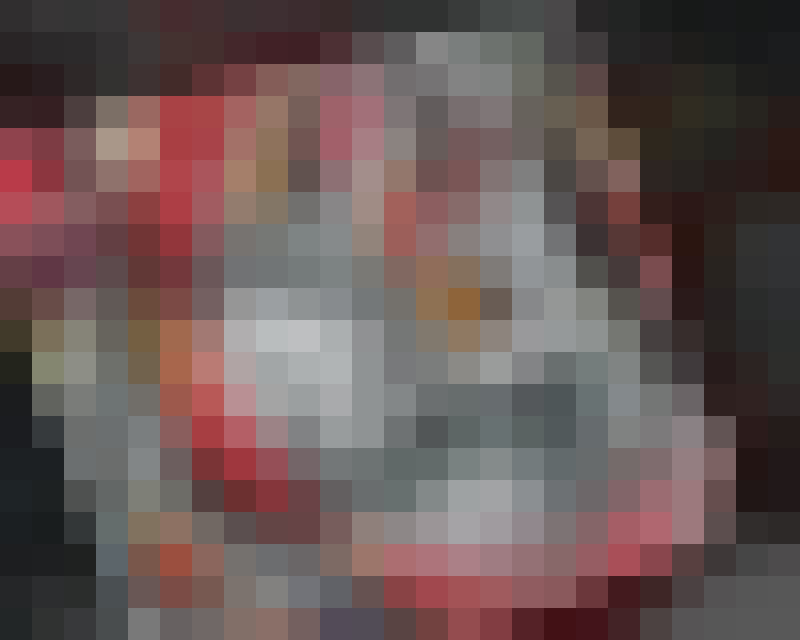} &
        \includegraphics[width=2.5cm]{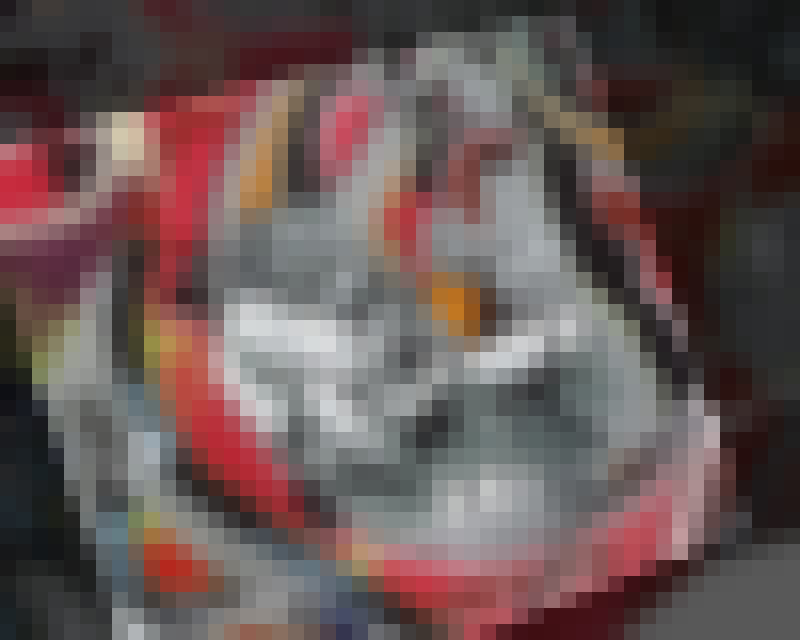} &
        \includegraphics[width=2.5cm]{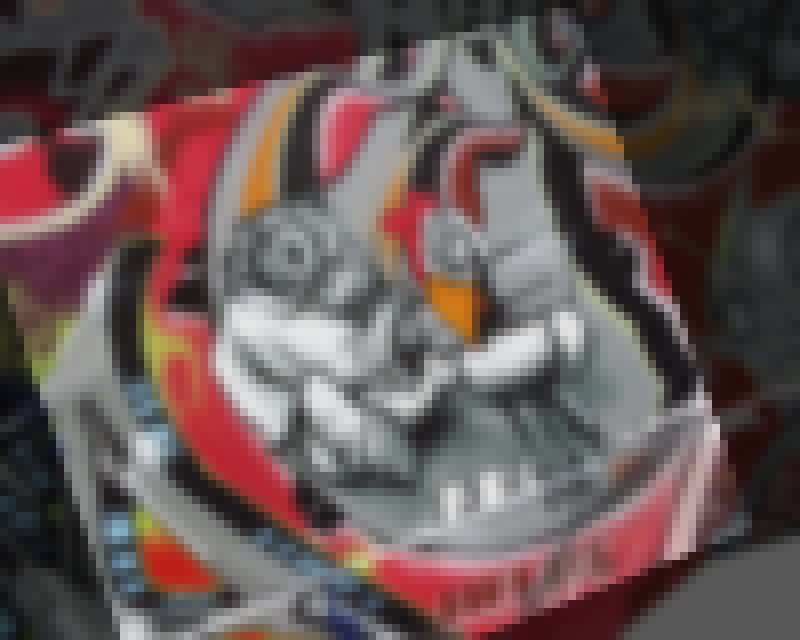} &
        \includegraphics[width=2.5cm]{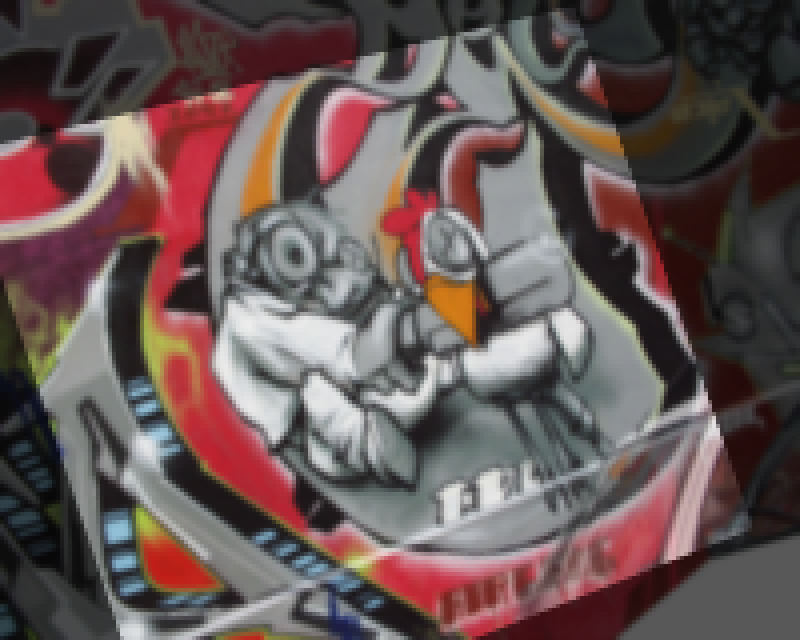} &
        \includegraphics[width=2.5cm]{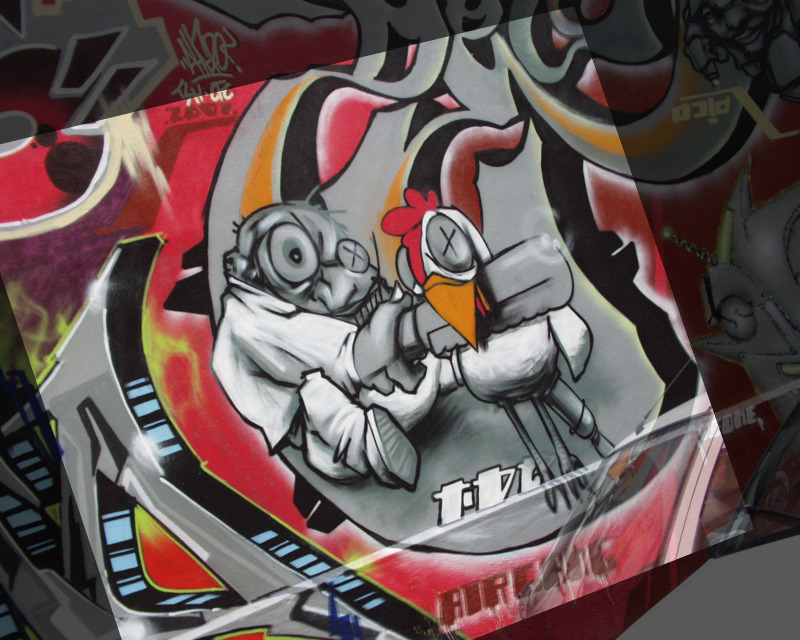} \\
        \includegraphics[width=2.5cm]{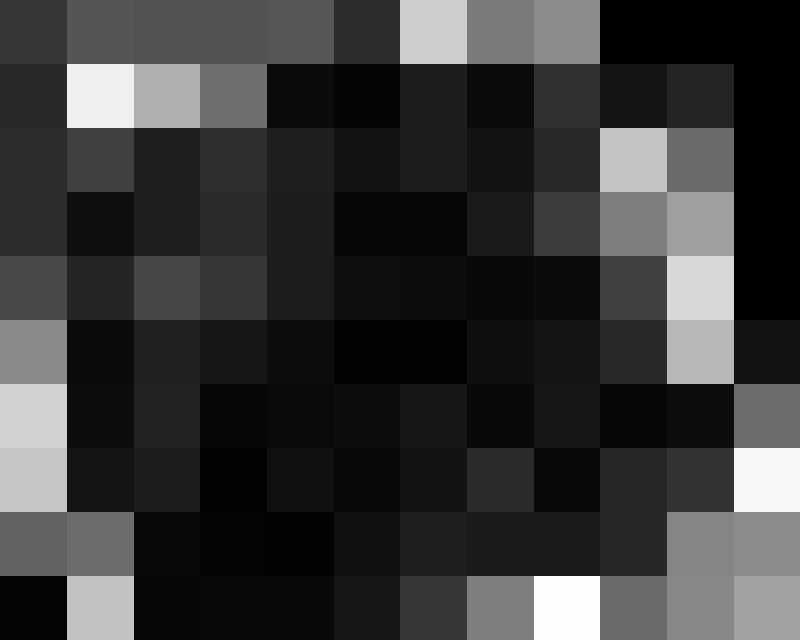} &
        \includegraphics[width=2.5cm]{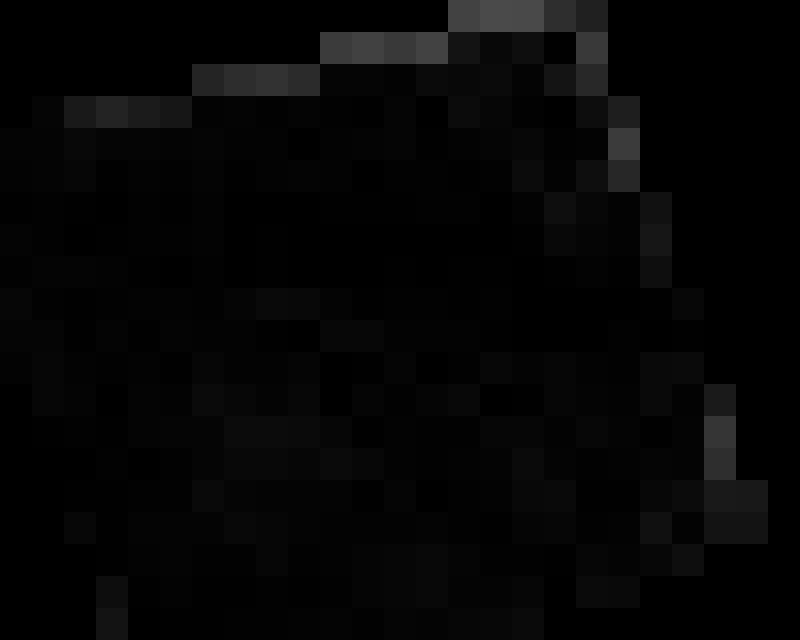} &
        \includegraphics[width=2.5cm]{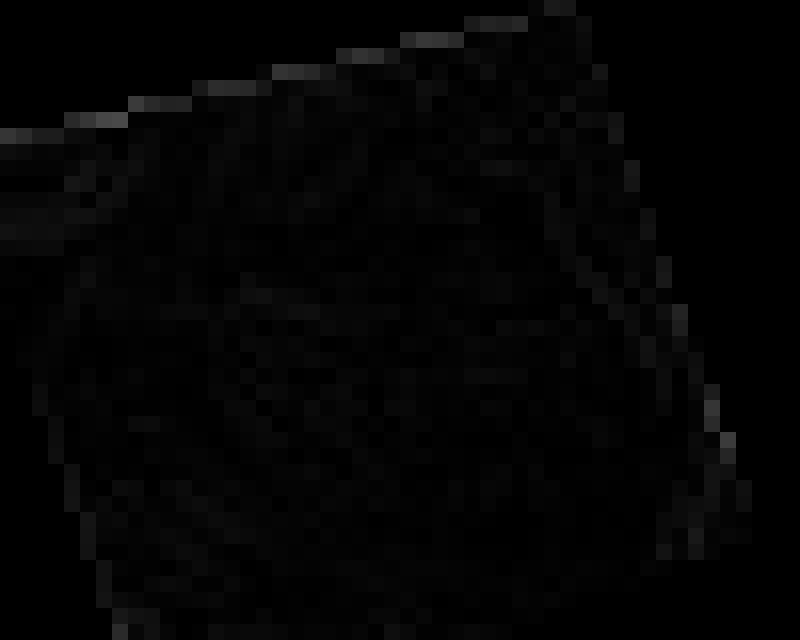} &
        \includegraphics[width=2.5cm]{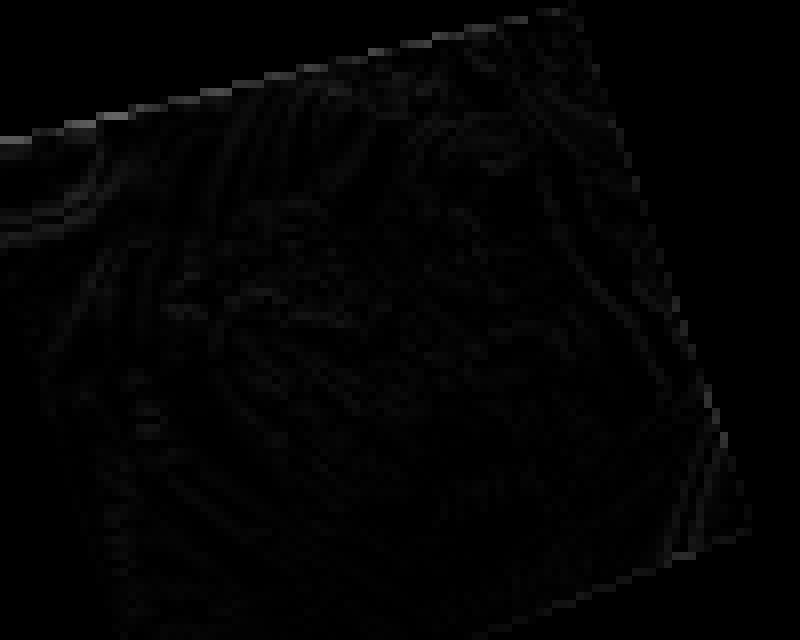} &
        \includegraphics[width=2.5cm]{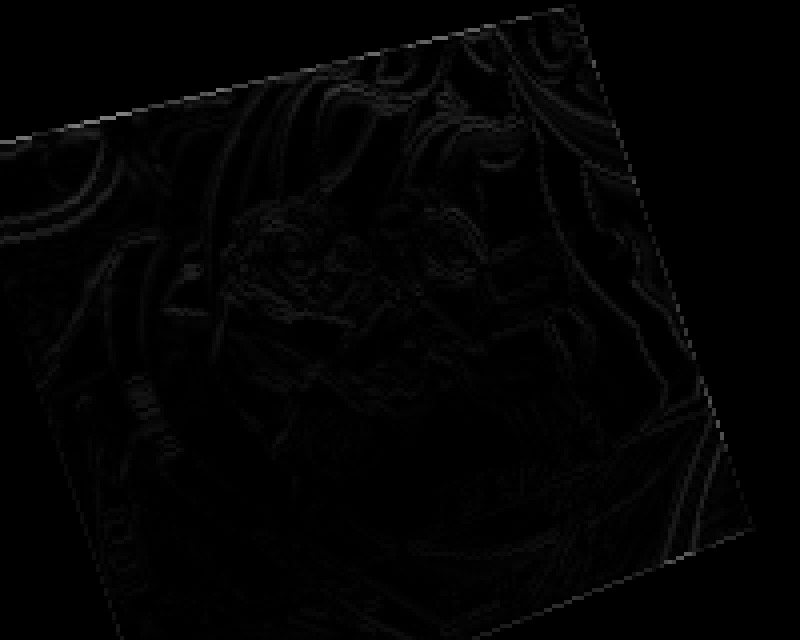} &
        \includegraphics[width=2.5cm]{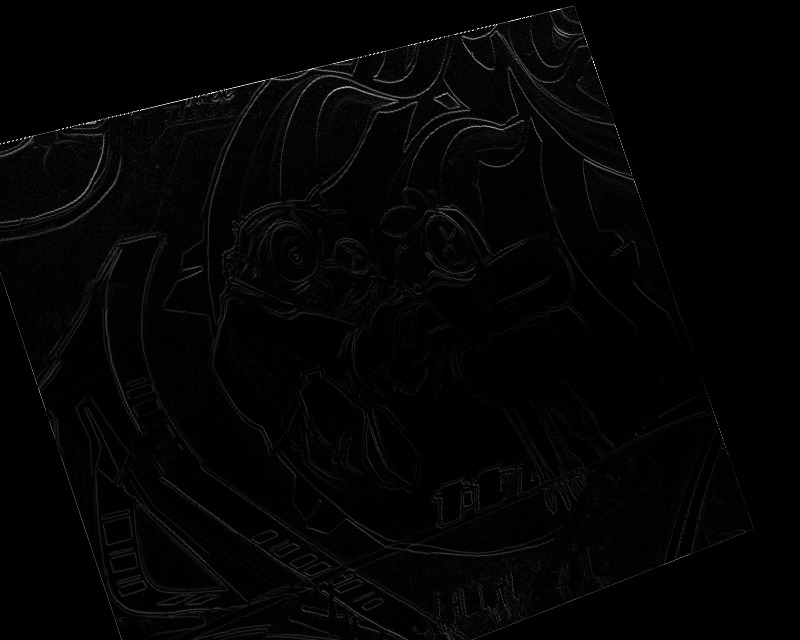} \\
        \end{tabular}
    \end{center}
    \caption{{\bf Results of the image registration by gradient descent.} Each  column represents a different level of the image pyramid used to optimize the loss function. \textit{Row 1:}  original source image; \textit{Row 2:}  original destination image; \textit{Row 3:}  source image warped to destination at the end of the optimization loop at that specific scale level.\textit{ Row 4:}  photometric error between the warped image using the estimated homography and the warped image using the ground truth homography. The algorithm starts to converge in the lower scales refining the solution as it goes to the upper levels of the pyramid.}
    \label{fig:homography_pyramid}
\end{figure*}

The loss function is optimized at each level of a multi-resolution pyramid, from the lower to the upper resolution levels. Figure \ref{fig:homography_pyramid} shows the original images, warped images and the error per pixel with respect to the ground truth warp at each of the scale levels. We use the Adam~\citep{adam2015} optimizer with a learning rate of $1e-3$, iterating 200 times at each scale level. As a side note, pixel coordinates are normalized in the range of $[-1, 1]$, meaning that there is no need to re-scale the optimized parameters between pyramid levels. The code for this example is provided in the following \underline{\color{blue}\href{https://github.com/kornia/kornia-examples/blob/master/homography.ipynb}{link}}.

\label{section:use_cases:depth_estimation}
\subsection{Multi-View Depth Estimation by Gradient Descent}
In this example we have implemented a fully differential generic multi-view pipeline, using our framework,  for machine learning research and applications. For this purpose, \lib{} provides the \mintinline{python}|kornia.DepthWarper| operator that takes an arbitrary number of calibrated camera views and warps them to a reference camera frame given the depth in the reference frame.

Multi-view reconstruction is a well understood problem with a good geometric model \citep{Hartley_MVG}, and many approaches for matching and optimization \citep{dtam_Newcombe2011, patchmatch-stereo, Sevilla-LaraSJB16} in addition to recent promising deep learning approaches \citep{Luo2016} making use of CNN's to extract robust features in order to create more accurate 3D reconstructions. We have found current machine learning approaches \citep{FischerDIHHGSCB15, IlgMSKDB16} to be limiting, in that they have not been generalized to arbitrary numbers of views (spatial or temporal); and available datasets \citep{Butler2012, Geiger2012CVPR} are only stereo and low resolution. Most machine learning approaches assume that there is high quality ground truth depth provided as commonly available datasets, which limits
their applicability to new datasets or new camera configurations. Classical approaches such as planesweep, patch match or DTAM~\citep{Newcombe:2011:DDT:2355573.2356447} have not been implemented with deep learning in mind, and do not fit easily into existing deep learning frameworks.\\

\begin{figure*}
    \setlength\tabcolsep{2.5pt}
    \begin{center}
        \begin{tabular}{c c c c c c c c}
        \textbf{Level 1} & \textbf{Level 2} & \textbf{Level 3} & \textbf{Level 4} & \textbf{Level 5} \\
        \includegraphics[width=2.7cm]{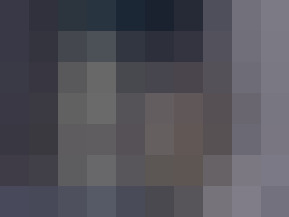} &
        \includegraphics[width=2.7cm]{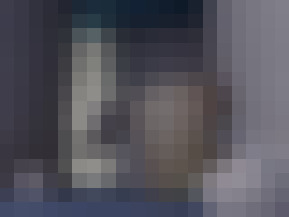} &
        \includegraphics[width=2.7cm]{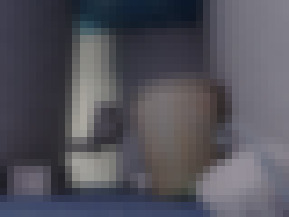} &
        \includegraphics[width=2.7cm]{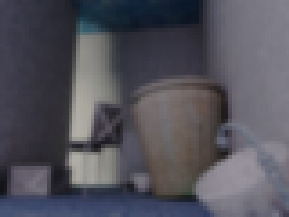} &
        \includegraphics[width=2.7cm]{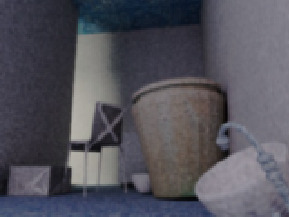} \\
        \includegraphics[width=2.7cm]{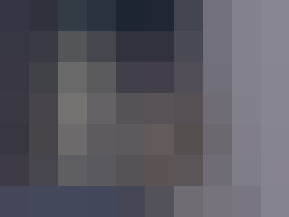} &
        \includegraphics[width=2.7cm]{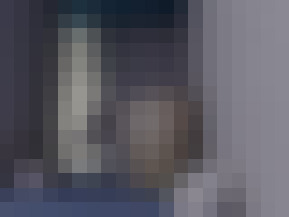} &
        \includegraphics[width=2.7cm]{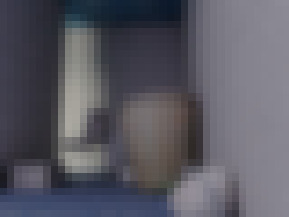} &
        \includegraphics[width=2.7cm]{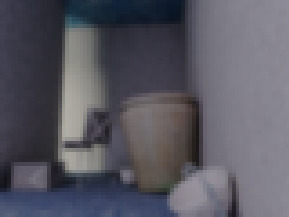} &
        \includegraphics[width=2.7cm]{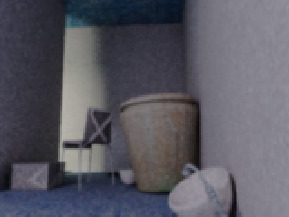} \\
        \includegraphics[width=2.7cm]{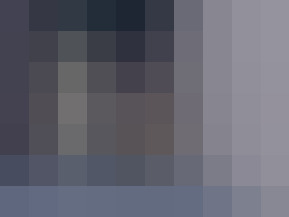} &
        \includegraphics[width=2.7cm]{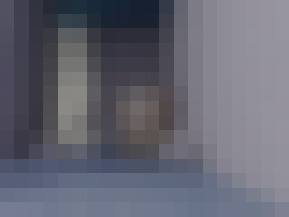} &
        \includegraphics[width=2.7cm]{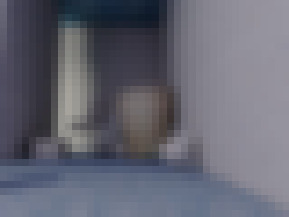} &
        \includegraphics[width=2.7cm]{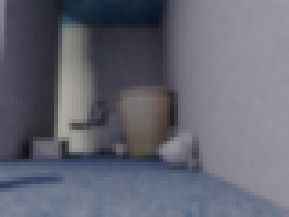} &
        \includegraphics[width=2.7cm]{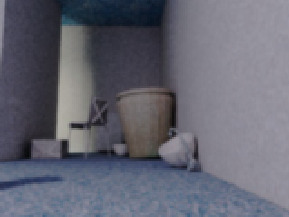} \\
        \includegraphics[width=2.7cm]{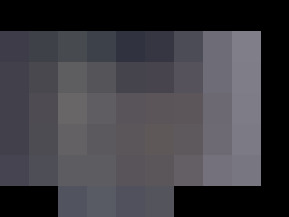} &
        \includegraphics[width=2.7cm]{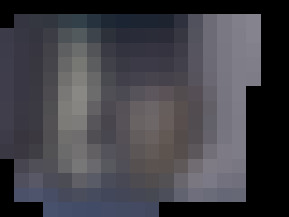} &
        \includegraphics[width=2.7cm]{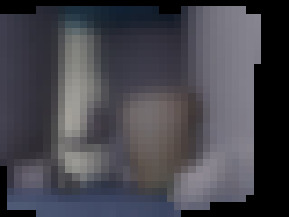} &
        \includegraphics[width=2.7cm]{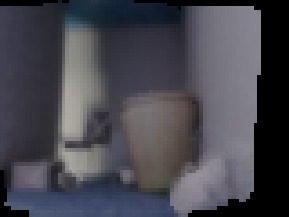} &
        \includegraphics[width=2.7cm]{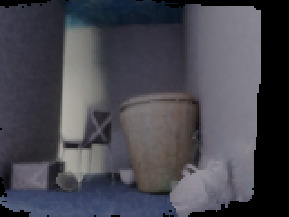} \\
        \includegraphics[width=2.7cm]{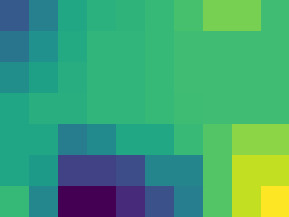} &
        \includegraphics[width=2.7cm]{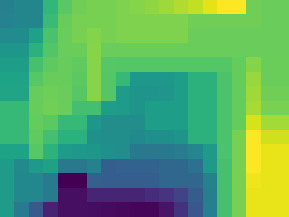} &
        \includegraphics[width=2.7cm]{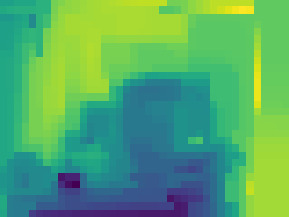} &
        \includegraphics[width=2.7cm]{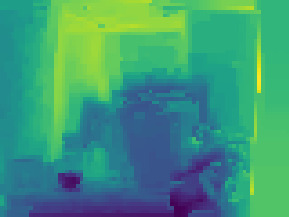} &
        \includegraphics[width=2.7cm]{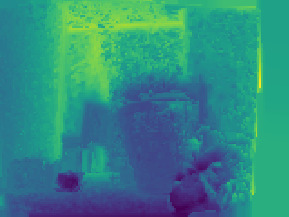} \\
        \includegraphics[width=2.7cm]{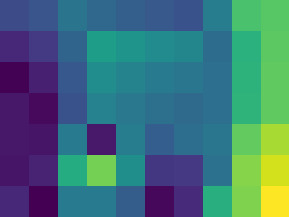} &
        \includegraphics[width=2.7cm]{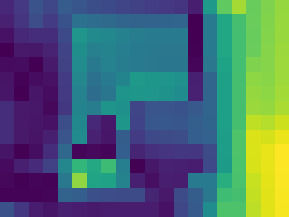} &
        \includegraphics[width=2.7cm]{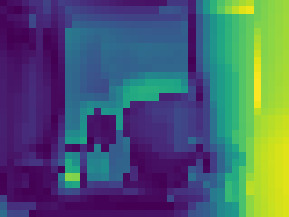} &
        \includegraphics[width=2.7cm]{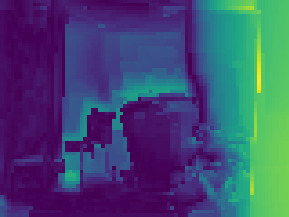} &
        \includegraphics[width=2.7cm]{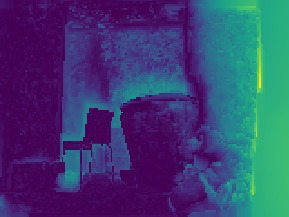} &
        \end{tabular}
    \end{center}
    \caption{{\bf Results of the depth estimation by gradient descent}. The results show the depth map produced by the given set of calibrated camera images over different scales. Each column represents a level of a multi-resolution image pyramid. \textit{Row 1 to 3:} source images, where the 2\textit{nd} row is the reference view; \textit{Row 4:} images from row 1 and 3 warped to the reference camera given the depth at that particular scale level. \textit{Rows 5 \& 6:} estimated depth map and the error per pixel compared to the ground truth depth map in the reference camera. The data used for these experiments was extracted from SceneNet RGB-D dataset \citep{McCormac:etal:ICCV2017}, containing photorealistic indoor image trajectories.}
    \label{fig:multiview:depth_estimation}
\end{figure*}

\begin{figure}[H]
    \begin{center}
        \includegraphics[width=0.45\linewidth]{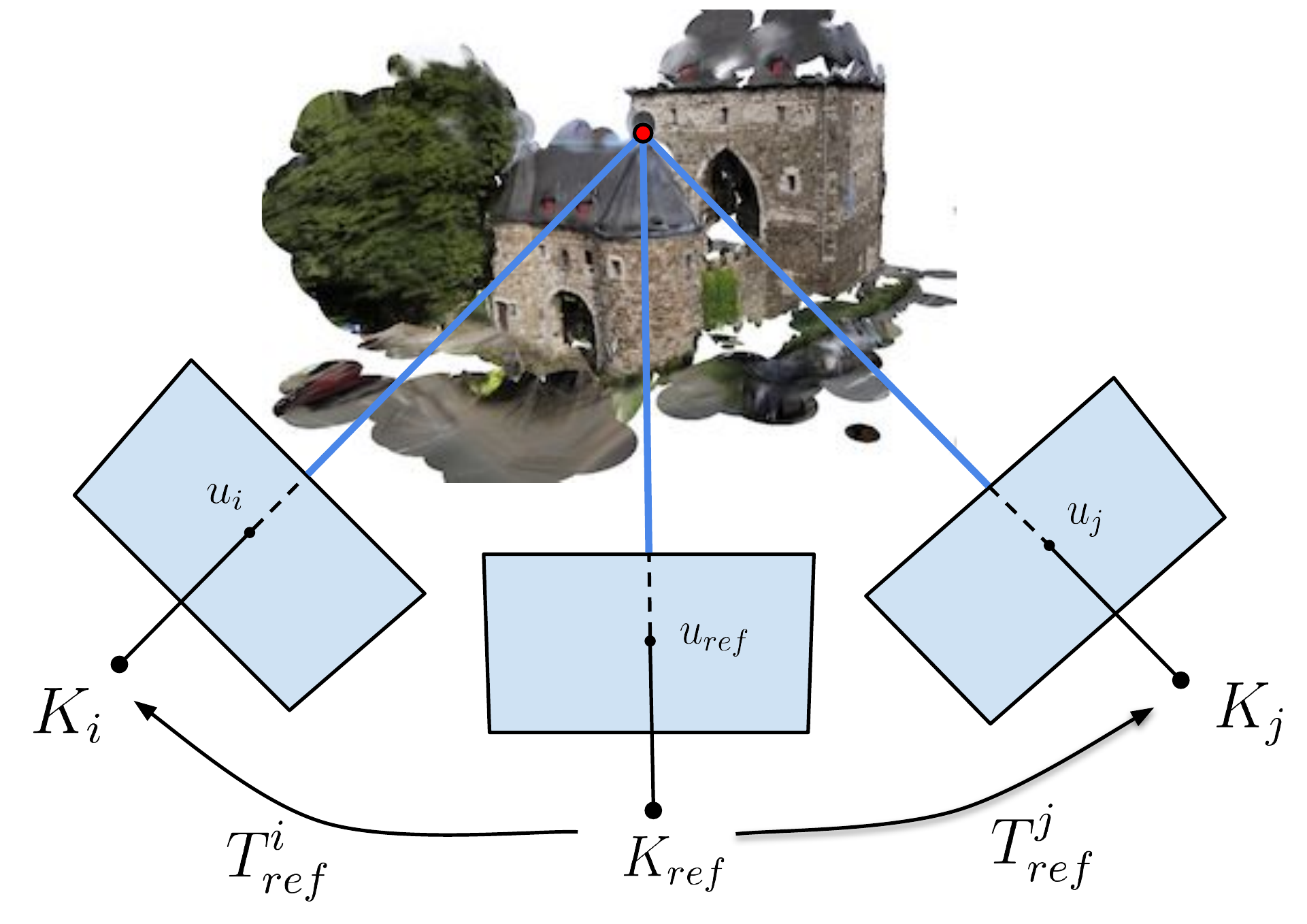} 
    \end{center}
    \caption{Classical multi-view stereo cameras setup with three cameras and the 2D point projections of a  3D point in the world reference frame. In the setup we describe in the text, we assume the internal parameters $K$ of the cameras are known, as well as their relative homogeneous transformations $T^{i}_{j}$. The subscript $_{ref}$ denotes the reference camera. } 
    \label{fig:depth:mvs-scheme}
\end{figure}

\textbf{Implementation.} We start with a simple formulation to estimate depth images using gradient descent on a variety of losses based on state of the art classical approaches (photometric, depth consistency, depth piece-wise smoothness, and multi-scale pyramids). The multi-view reconstruction pipeline receives as input a set of views, including RGB images and calibrated intrinsic camera models $K_{i}$ and relative pose estimates $T_{\text{ref}}^{i}$.  Then, the system   estimates the depth image $\boldsymbol{d}_{\text{ref}}$ for a reference view. Since we assume a calibrated setup, the depth value of a given pixel $\boldsymbol{u}_{\text{ref}} = [u_{\text{ref}},v_{\text{ref}}]$ in the reference view, $\boldsymbol{d}_{\text{ref}}$, can be used to compute the corresponding pixel location $\boldsymbol{u}_i = [u_{i},v_{i}]$ in any of the other views through simple projective geometry $H_{\text{ref}}^{i} = K_i \cdot T_{\text{ref}}^{i} \cdot K_{\text{ref}}^{-1}$. Given this, we can warp views onto each other 
using a differentiable bilinear sampling, as proposed in~\citep{stn_NIPS2015}, $\tilde{I_{\text{ref}}} = \omega(I_i, H_{\text{ref}}^{i}, \boldsymbol{d}_{\text{ref}})$.

Similar to \citep{dtam_Newcombe2011, monodepth17, superdepth18}, depth is estimated by minimizing a photometric error between the views warped to the reference view:
\begin{align}
	\label{eq:multiview:loss photometric1}
	L_{\text{photo1}} &= \frac{1}{n} \sum\limits^{n} \dfrac{1 - \text{SSIM}(I_{\text{ref}}, \tilde{I_{\text{ref}}})}{2}
	\\
	\label{eq:multiview:loss photometric2}
	L_{\text{photo2}} &= \frac{1}{n} \sum\limits^{n} \lvert I_{\text{ref}} - \tilde{I_{\text{ref}}}\rvert
\end{align}

We compute an additional loss to encourage disparities to be locally smooth with a penalty on the disparity gradients weighted by image gradients as 

\begin{equation}
	\label{eq:multiview:loss smoothness}	
	L_{\text{smooth}} = \frac{1}{n} \sum\limits^{n} \lvert \partial_{x} \, d \rvert e^{- \| \partial_{x} I_{i} \|} + \lvert \partial_{y} \, d \rvert e^{- \| \partial_{y} I_{i} \|}
\end{equation}

Finally, losses are combined with a weighted sum:

\begin{equation}
   	\label{eq:multiview:loss total}
	L_{\text{total}} = \alpha L_{\text{photo1}} + (1 - \alpha) L_{\text{photo2}} + \lambda L_{\text{smooth}} 
\end{equation}
These losses could be easily modified or extended, depending on how well their inherent  assumption fit the data, e.g. photometric consistency only holds for small view displacements.


Figure~\ref{fig:multiview:depth_estimation} shows partial results obtained by the depth algorithm implemented using \lib. The algorithm receives as input 3 calibrated RGB images (320x240). We used Stochastic Gradient Descent (SGD) with momentum and compute the depth at 7 different scales by blurring the image and down-sampling the resolution by a factor of 2 from the previous size. To compute the loss, we up-sample  to the original size using bilinear interpolation.  The refinement at each level was done for 500 iterations starting from the lowest resolution. The initial values for depth were obtained by a random uniform sampling in a range between 0 and 1. The code for this example is provided in the following \underline{\color{blue}\href{https://github.com/kornia/kornia-examples/blob/master/depth_estimation.ipynb}{link}}.

\subsection{Targeted adversarial attack on SIFT-matching}
\label{section:use_cases:adversarial_matching}

\begin{figure}[t]
    \begin{center}
        \includegraphics[width=0.95\linewidth]{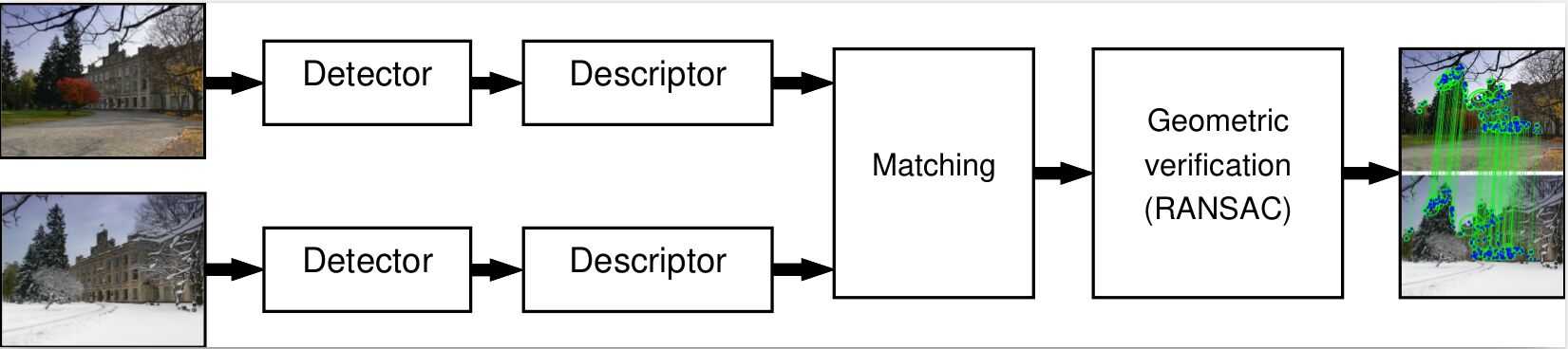} 
    \end{center}
    \caption{The diagram of the commonly used wide baseline stereo matching algorithm~\citep{Pritchett1998}.}
    \label{fig:wbs-scheme}
\end{figure}

\begin{figure*}[tb]
    \begin{center}
        \includegraphics[width=0.32\linewidth]{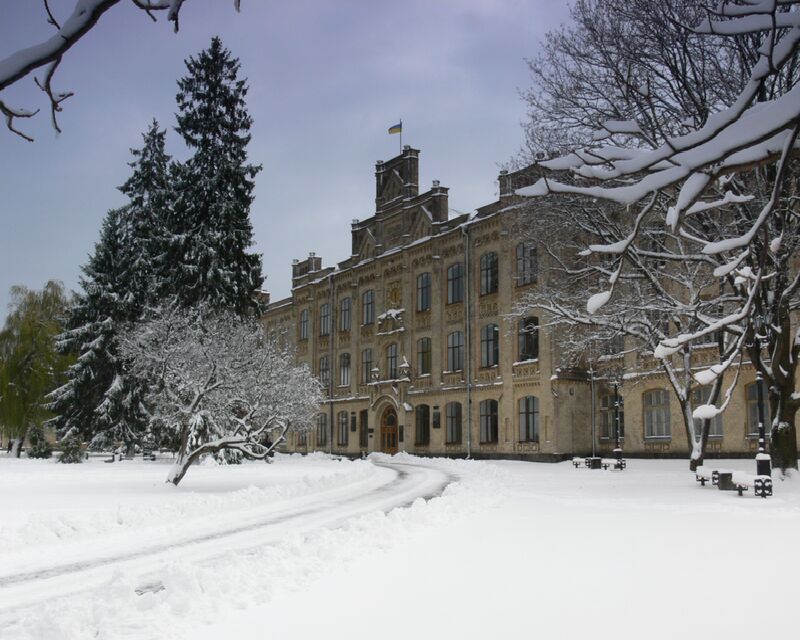}    \includegraphics[width=0.32\linewidth]{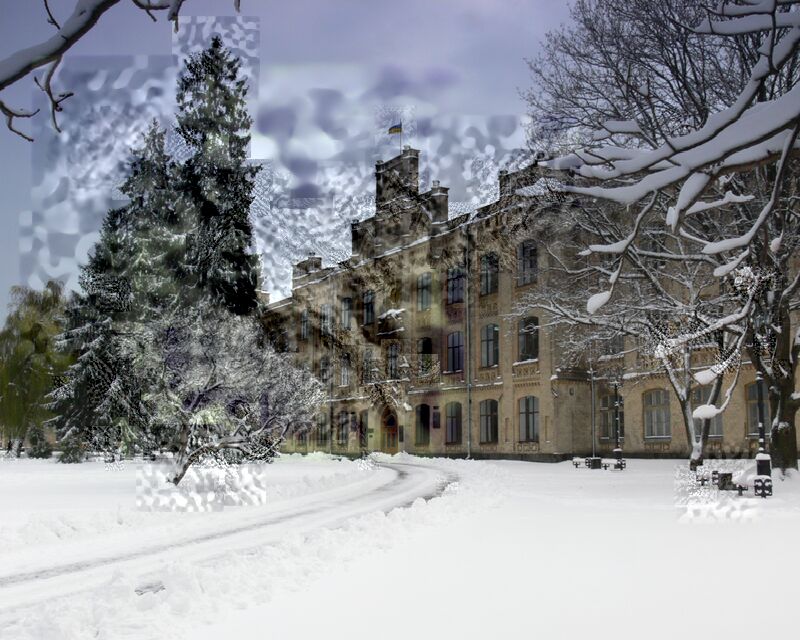} 
        \includegraphics[width=0.32\linewidth]{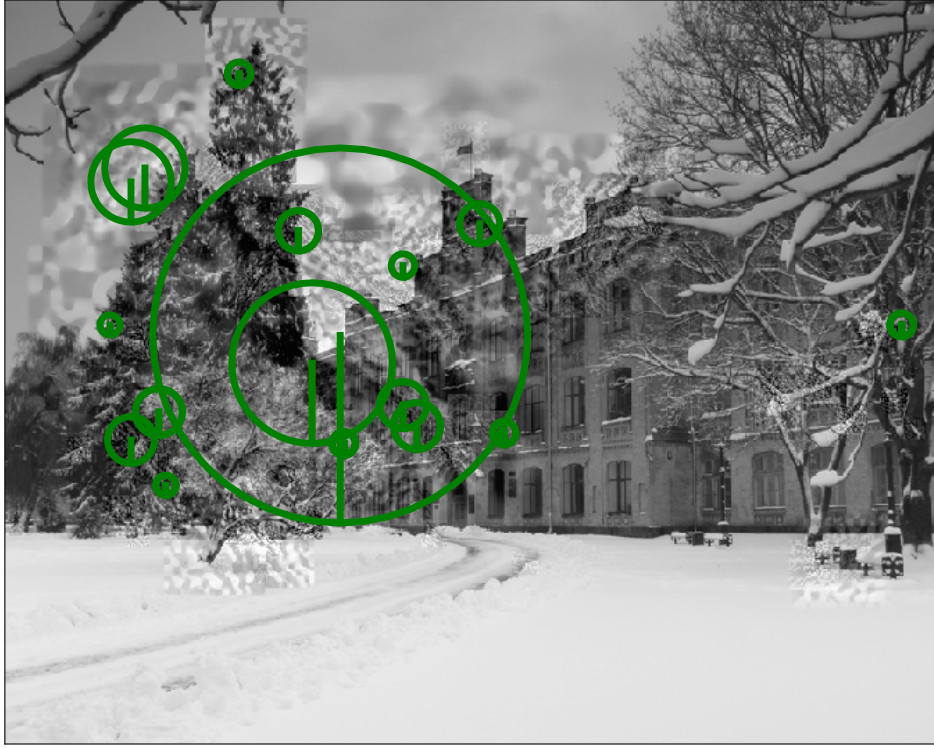} \\
        \includegraphics[width=0.32\linewidth]{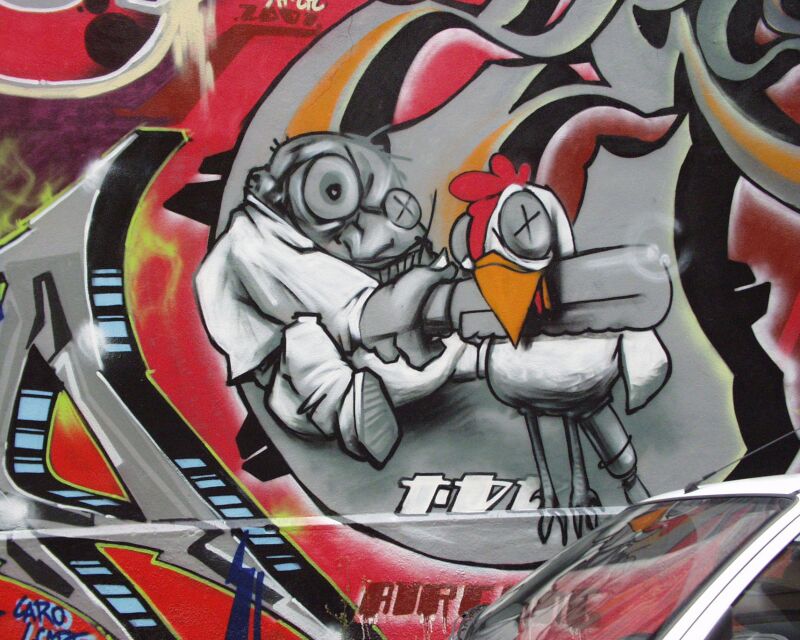} 
        \includegraphics[width=0.32\linewidth]{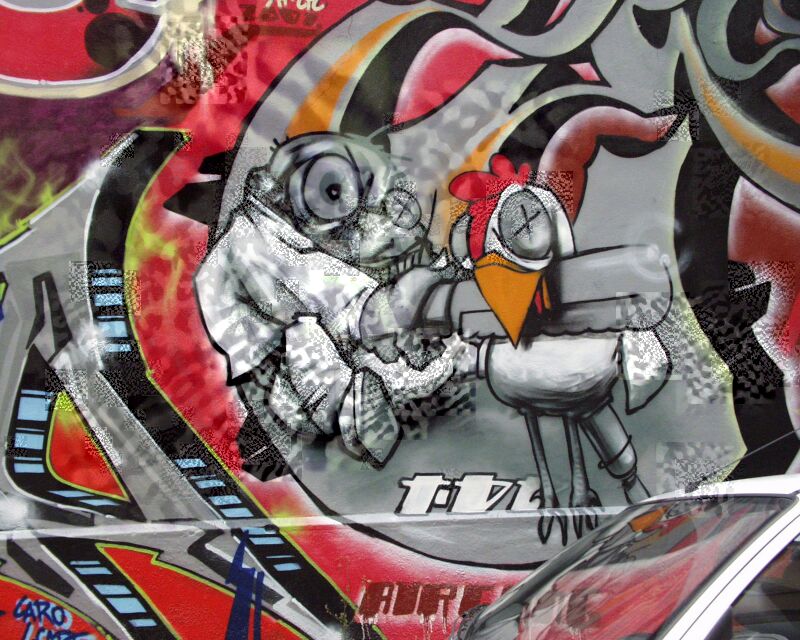} 
        \includegraphics[width=0.32\linewidth]{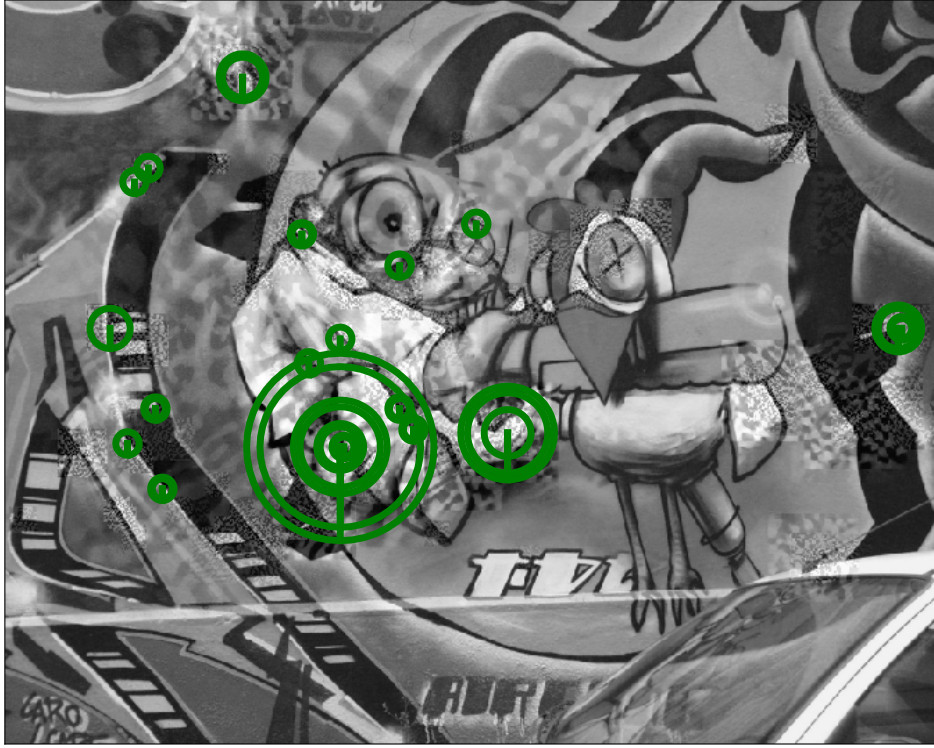} 
    \end{center}
    \caption{{\bf Targeted adversarial attack on image matching.} From left to right: original images, which do not match; images, optimized by gradient descent to have local features that match; the result of the attack: matching features (Hessian detector + SIFT descriptor), which survived RANSAC geometric verification}
    \label{fig:wbs}
\end{figure*}
In this final  example we show how to implement a  fully differential wide baseline stereo matching with local feature detectors and descriptors (see Figure~\ref{fig:wbs-scheme}) using \mintinline{python}|kornia.features|. We demonstrate the differentiability by making a targeted adversarial attack on the wide baseline matching pipeline.

\textbf{Local feature detectors and descriptors} are the workhorses of 3d reconstruction~\citep{schonberger2016structure, torii2018structure}, visual localization~\citep{sarlin2019coarse} and image retrieval \citep{shen2018matchable}. Although learning-based methods now seem to dominate~\citep{LocaFeaturesReview2018}, recent benchmark top-performers still use Difference-of-Gaussians aka SIFT detector~\citep{CVPRW2019}. SIFT descriptor is still one of the best for 3d reconstruction tasks~\citep{ColmapBenchmark2017}. Thus, we believe that community would benefit from having GPU-accelerated and differentiable version of the classical tools provided under \mintinline{python}|kornia.features|.

\textbf{Adversarial attacks.}
Adversarial attacks is an area of research which recently gained popularity after the seminal work of Szegedy et al.~\citep{AdvAttack2014}, showing that small perturbations in the input image can switch the neural network prediction outcome. There exist several  works showing that CNN-based solutions for classification~\citep{NIPS2018Adv}, segmentation~\citep{arnab_cvpr_2018}, object detection~\citep{ObjDetAdv2018} and image retrieval~\citep{AdvRetrieval2019}   are all prone to such attacks. 
The only paper about attack on local feature matching is~\citep{adversarial_attack_local_features2019l}, which proposed to place special noisy patches on response peak locations, killing the matching process for matching pairs. Yet, the authors do not know of any paper devoted to targeted adversarial attacks on local features-based image matching. 
Most attack methods are "white-box"~\citep{NIPS2018Adv}, which means they require access to the model gradients w.r.t the input. This makes them an excellent choice for a \mintinline{python}|kornia.features| differentiability demonstration.

\textbf{Implementation.}
The two view matching task is posed as follows~\citep{Pritchett1998}: given two images $I_{a}$ and $I_{b}$ of the same scene, find the correspondences between pixels in images. This is typically solved by detecting local features, describing the local patches with a descriptor and then matching by minimum descriptor distance with some filtering. \lib{} provides all these ingredients. 

We consider the following adversarial attack: given the non-matching image pair $I_{a}$, $I_{b}$, and the desired homography $H_a^b$, modify images so that the correspondence finding algorithm will output a non-negligible number of matches consistent with the homography $H_a^b$. This means that both local detectors should fire in specific locations and the local patches around that location should be matchable by a given loss function such as:

\begin{align}
	\label{eq:adversarial:loss total}
	L_{\text{total}} &= L_{\text{loc}} + \alpha L_{\text{desc}} + \beta L_{\text{reg}}\\
	\label{eq:adversarial:loss loc}
	L_{\text{loc}} &= \frac{1}{n} \sum\limits^{n} (p_1 - H p_2)^2 \\
   \label{eq:adversarial:loss desc}
	L_{\text{desc}} &= \frac{1}{n} \sum\limits^{n} (1 + d(D_1, D_2) - d(D_1, D_{2neg})) \\
	\label{eq:adversarial:loss reg}
	L_{\text{reg}} &= \frac{1}{n} \sum\limits^{n} (I - I_{\text{init}})^2 
\end{align}

\noindent where $p_1$ is keypoint detected in $I_a$, $p_2$ is closest reprojected by the $H_a^b$ keypoint detected in image $I_b$, $\sigma_1$ and $\sigma_2$ are their scales, $D_1$ and $D_2$ -- their descriptors, $D_{2neg}$ - hard negative in batch, $d(\cdot, \cdot)$ -- L2 distance, and $I_{\text{init}}$ is original unmodified version of $I_a$ and $I_b$.

The detector used in this example is the Hessian blob detector~\citep{Hessian78}; the descriptor is the SIFT~\citep{Lowe2004}. We keep the top-2500 keypoints and use the Adam~\citep{adam2015} optimizer with a learning rate of 0.003. 
Figure~\ref{fig:wbs} shows the original images, optimized images and optimized images with matching features visualized.  The perturbations are not quite imperceptible, but that it is not the goal of the current example. The code for this example is provided in the following \underline{\color{blue}\href{https://github.com/kornia/kornia-examples/blob/master/local-feature-adversarial-attack.ipynb}{link}}.

\section{Conclusions}
We have introduced \lib, a library for computer vision in PyTorch that implements traditional vision algorithms in a differentiable manner making use of hardware acceleration to improve performance. We demonstrated that using our library, classical vision problems such as image registration by homography, depth estimation, or local features matching can be very easily solved with a high performance similar to existing libraries making use of GPU acceleration and batch parallelism. By leveraging this project, we believe that classical computer vision libraries can take a different role within the deep learning environments as components of layers of the networks as well as pre- and post-processing of the results. In addition, one more reason why combining traditional and deep learning methods through \lib{} is the fact that it can help to minimize the time-cost of having engineers tuning hyper-parameters for classical vision methods where instead those parameters can be easily optimised for specific tasks making more powerful the use of differentiability. In the future, we expect researchers and companies increase the number of contributions to the project to grow on other areas such as Structure from Motion or volumetric image transformations. At the time of this submission, \lib{} has 2450 GitHub stars and 250 forks and up to 50 contributors and it's been used by more than 50 projects.

\section*{Acknowledgements}

We would like to acknowledge Arraiy, Inc. to sponsor the initial stage of the project. The folks from the Open Source Vision Foundation and OpenCV.org, and the PyTorch open-source community for helpful contributions and feedback. The work of Edgar Riba and Daniel Ponsa has been partially supported by the Spanish Government under Project TIN2017-89723-P, and the work of Francesc Moreno-Noguer by projects HuMoUR TIN2017-90086-R and ERA-Net Chistera IPALM PCI2019-103386. Dmytro Mishkin is supported by CTU student grant SGS17/185/OHK3/3T/13 and by supported by the Austrian Ministry for Transport, Innovation and Technology, the Federal Ministry for Digital and Economic Affairs, and the Province of Upper Austria in the frame of the COMET center SCCH.

\bibliographystyle{unsrt}
\bibliography{main}  






\end{document}